\documentclass[final,5p,times,twocolumn,authoryear]{elsarticle}

\usepackage{amssymb}
\usepackage{amsmath}

\usepackage{algorithm}
\usepackage{algpseudocode}
\usepackage{array}
\usepackage[caption=false,font=normalsize,labelfont=sf,textfont=sf]{subfig}
\usepackage{textcomp}
\usepackage{stfloats}
\usepackage{url}
\usepackage{verbatim}
\usepackage{graphicx}
\usepackage{multirow}
\usepackage{booktabs}
\usepackage{tabularx}
\usepackage{xcolor}
\usepackage{float}
\usepackage{placeins}

\newcolumntype{Y}{>{\raggedright\arraybackslash}X}
\newcommand{\tablefontsize}{\footnotesize}

\journal{Expert Systems with Applications}

\begin{document}

\begin{frontmatter}

\title{Coarse Semantic Injection for LLM-Conditioned Structured Indoor Prediction}

\author{Shuliang Zhu, Tomiwa Adey, Jinjia Zhou}

\begin{abstract}
Large language models (LLMs) have recently been used as structured decoders for indoor understanding from 3D point-token inputs. However, point cloud encoders often under-represent thin structural elements such as doors and windows after voxelization and sparse pooling, and may miss individual furniture instances in cluttered scenes. We propose an interface-preserving semantic augmentation for LLM-conditioned structured decoding. The key idea is to associate semantic evidence with the point-cloud representation, reduce it to a coarse four-group code (furniture, walls, openings, and others), and encode it as an RGBB point interface: red for furniture, green for walls, blue for openings, and black for others, where RGBB denotes four semantic color states represented in three RGB channels rather than an additional fourth channel. This semantic color code is appended to the original raw point attributes before tokenization, so geometry and semantics share the same sparse tokenization path while the downstream language model decoder and output serialization remain unchanged. We further introduce a lightweight routed semantic shift module, with an auxiliary head used only for training-time ratio/budget regularization and analysis, to strengthen semantic cues after sparse pooling. The overall pipeline can use RGB-derived semantic evidence. Under these controlled semantic-source settings, the reported metrics improve across Structured3D, the SpatialLM dataset, and ARKitScenes, especially for opening localization and per-instance furniture detection in cluttered scenes. Ablations clarify the roles of semantic source, color coding, token fusion, and shift injection, while also showing that color/entropy effects remain nontrivial. 
\end{abstract}

\begin{keyword}
Structured indoor prediction \sep large language models \sep point cloud tokenization \sep semantic injection
\end{keyword}

\end{frontmatter}


\section{Introduction}

Structured indoor modeling aims to recover a compact and parameterized representation of indoor scenes, typically including architectural layout elements such as walls, doors, and windows, as well as object-level 3D bounding boxes. Such structured outputs support downstream applications including robotic navigation, embodied interaction, AR content authoring, and editable indoor digital twins. Early indoor parsing and layout estimation methods studied geometric room regularities from images~\cite{hedau2009recovering,lee2009geometric}, while later floorplan and layout reconstruction methods improved structured prediction from panoramic or multi-view observations~\cite{zou2018layoutnet,sun2019horizonnet,yue2023roomformer}. Classical 3D systems reconstruct scene geometry through structure-from-motion, multi-view stereo, and optimization-based fitting pipelines~\cite{schonberger2016structure}, whereas learning-based methods increasingly predict dense geometry, layout, or object structure directly from images, posed video, or point representations~\cite{nie2020total3d,murez2020atlas,sun2021neuralrecon}. More recently, autoregressive and language-conditioned models have been explored as structured decoders for indoor scenes, enabling walls, openings, and object boxes to be represented as serialized outputs rather than as independent task heads~\cite{avetisyan2024scenescript,mao2025spatiallm,hong20233dllm}.

\begin{figure}[t]
\centerline{\includegraphics[width=0.5\textwidth, clip, trim=0cm 3.2cm 0cm 2.8cm]{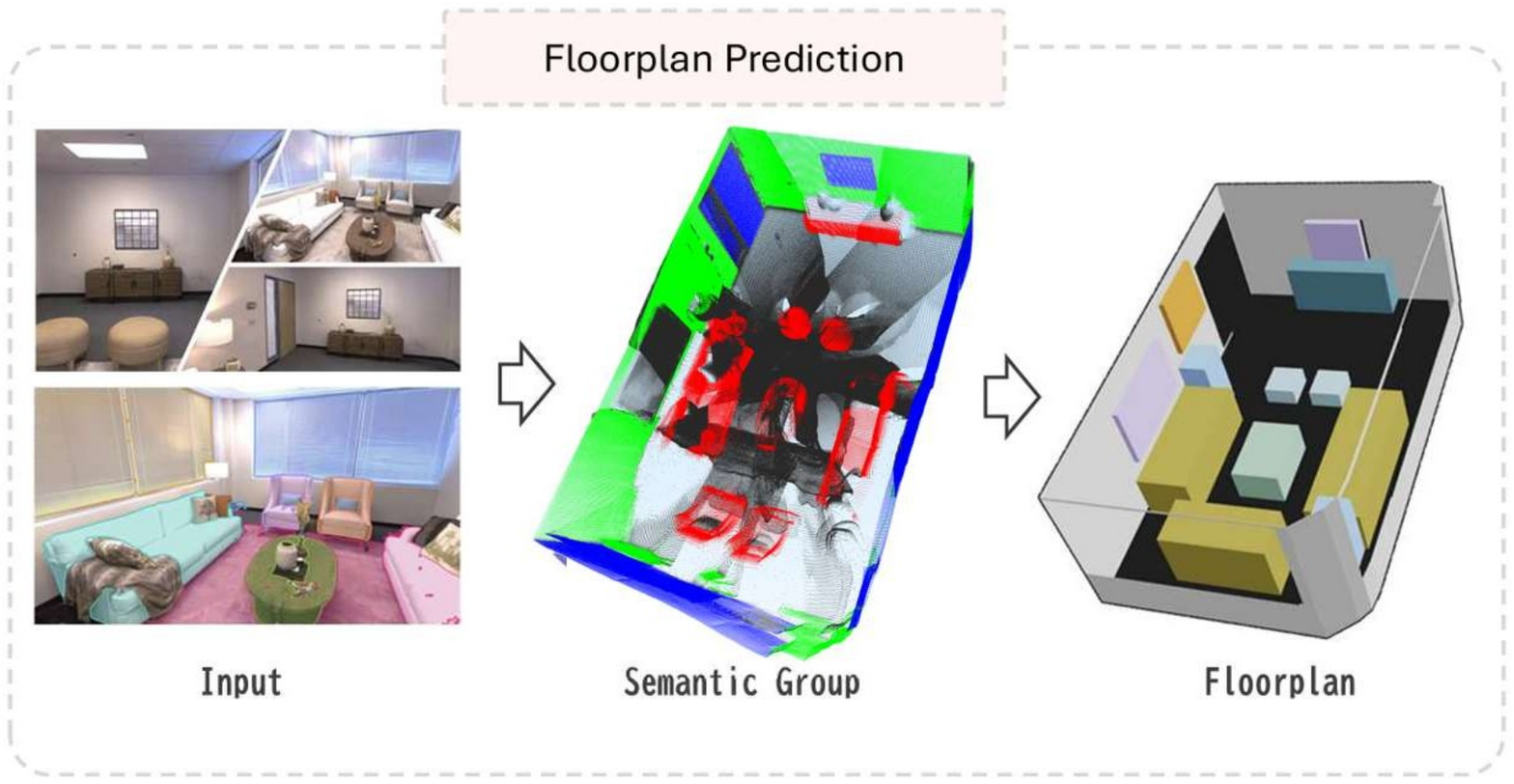}}
\caption{Structured indoor modeling with semantic-colored point conditioning. RGB views can provide semantic evidence, which is associated with the point-cloud representation and decoded into structured layout/floorplan outputs. In the controlled experiments, evaluation starts from the point-cloud interface to isolate the effect of semantic-colored conditioning.}
\label{fig:teaser}
\end{figure}

In the LLM-conditioned paradigm, point cloud encoders convert 3D geometry into compact token sequences through voxelization, sparse serialization, and pooling~\cite{qi2017pointnet,qi2017pointnetplusplus,thomas2019kpconv,zhao2021pointtransformer,qian2022pointnext,wu2025sonata,mao2025spatiallm}. This design provides an efficient bridge between unordered 3D points and sequence decoders, and recent self-supervised or masked point pretraining further improves the transferability of point representations~\cite{yu2022point,pang2022pointmae,wu2025sonata,zhang2026utonia}. However, sparse point-token pipelines still tend to attenuate fine structural evidence. Two failure modes are particularly common in structured indoor prediction: (i) thin or boundary-dominated elements such as doors and windows, which occupy few voxels and whose geometric evidence can be diluted during sparse pooling, and (ii) individual furniture instances in cluttered multi-object arrangements, where overlapping spatial extents reduce per-instance discriminability and frequently lead to missed boxes. These errors are related to broader challenges observed in 3D detection and point-based scene understanding, where sparse support, occlusion, and clutter can strongly affect object recall~\cite{qi2019votenet,liu2021groupfree3d,choy20194d}. We address this limitation by modifying the point-token conditioning interface rather than replacing the downstream language decoder or changing the serialized output protocol.

A natural complementary signal is coarse semantics derived from RGB views. Modern image segmentation models provide strong object and region masks from 2D images, ranging from fully convolutional semantic segmentation to instance, panoptic, and promptable segmentation~\cite{long2015fcn,he2017maskrcnn,cheng2021maskformer,cheng2022mask2former,kirillov2023sam,carion2025sam}. Such cues are especially relevant for the categories where point cloud encoders struggle: doors and windows are often visually salient in 2D even when their 3D support is thin, and individual furniture instances can be separated in the image domain even when their 3D extents overlap. Related open-vocabulary and multimodal 3D mapping works also show that lifting 2D semantic or language-aligned features into 3D can improve scene-level interpretability and object-level association~\cite{radford2021learning,jatavallabhula2023conceptfusion,peng2023openscene,takmaz2023openmask3d,kerr2023lerf,shafiullah2023clipfields}. In contrast to these works, our goal is not open-vocabulary 3D segmentation or semantic mapping itself, but a minimal semantic conditioning interface for LLM-conditioned structured indoor decoding.

Accordingly, we focus on a deliberately compact semantic representation. After lifting 2D masks to 3D, we partition points into four groups that are easy to obtain and directly relevant to structured indoor parsing: furniture, walls, openings (doors/windows), and ``others''. We encode these groups using a fixed four-color code in RGB space: \textbf{R} for furniture, \textbf{G} for walls, \textbf{B} for openings, and \textbf{black} for other points. This design provides point-level semantic association without requiring dense multi-class semantic supervision or an additional structured prediction head. It is also compatible with the same point-tokenization backbone and autoregressive language decoder used by SpatialLM-style pipelines~\cite{mao2025spatiallm,wu2025sonata,wang2024qwen2}.

The main technical challenge is to make such coarse 2D semantics survive the 3D tokenization pipeline. Voxelization, serialization, and sparse pooling can attenuate thin semantic regions and smear object boundaries, especially when the semantic evidence is injected only after encoding. Moreover, token-level fusion that assumes correspondence between geometric tokens and semantic tokens is brittle, because pooling and ordering are not guaranteed to align across independently processed streams. Conditioning mechanisms such as feature-wise modulation have shown that auxiliary signals can effectively steer visual representations~\cite{de2017modulating,perez2018film}, but structured indoor prediction requires the conditioning signal to remain spatially localized for doors, windows, and individual furniture instances. Therefore, simply appending global semantic features or fusing separate post-encoder streams is insufficient for our setting.

We address this by designing a tokenization-compatible semantic interface: the four-color semantic code is appended to the original raw point attributes before sparse tokenization, so semantic cues and geometry share the same serialization and pooling path. On top of the resulting token sequence, we apply a lightweight semantic shift module that performs class-conditional residual injection with a geometry-side router. This module strengthens semantic evidence after sparse pooling while preserving the original LLM decoder and output serialization. The intervention is intentionally small: it improves the conditioning interface rather than introducing a new task-specific detector, layout head, or segmentation decoder.

Beyond average-case improvements, the semantic color interface also provides input-level controllability. In the practical branch, a user can click on a target object in a 2D view, obtain a promptable segmentation mask~\cite{kirillov2023sam,carion2025sam}, associate the mask with the point-cloud representation, and replace the original raw point-cloud input with a target-specific semantic-colored input: only the target support points are activated and all other points are black. Thus, the paper evaluates both a controlled point-cloud-interface setting and a practical click/SAM controllability setting within the same overall pipeline.

We evaluate the proposed method primarily at the point-cloud interface. On Structured3D and the SpatialLM dataset~\cite{zheng2020structured3d,mao2025spatiallm}, the language model decoder autoregressively generates a serialized structured description of layout elements and furniture boxes from point tokens. On ARKitScenes~\cite{baruch2021arkitscenes}, the same decoder predicts oriented furniture boxes from fixed point-cloud inputs, while the click/SAM study evaluates the RGB/SAM mask-acquisition branch for controllability. With only 0.2M additional trainable parameters over the SpatialLM baseline, the proposed interface improves controlled metrics for openings and furniture while preserving the structured decoding interface.

\paragraph{Contributions}
Our contributions are four-fold:
(i) We formulate an \textbf{interface-preserving semantic-colored point conditioning mechanism} that encodes coarse four-group semantics (furniture / walls / openings / others) as a fixed four-color point interface (R/G/B/black) and injects it before sparse tokenization so that semantic cues traverse the point-token pipeline.
(ii) We introduce a \textbf{class-conditional routed semantic shift injection} with only 0.2M additional trainable parameters, together with ratio/budget/entropy terms used as lightweight training-time regularizers and analysis tools. These components are intended to reduce degenerate routing and expose the induced layout--box trade-off, rather than to define a separate stability-learning paradigm.
(iii) We establish a \textbf{controlled evaluation protocol over semantic sources and interface variants}, including oracle RGBB, random four-color, post-encoder fusion, and no-shift variants, to diagnose when semantic coloring helps and where color/entropy confounds remain.
(iv) We provide a \textbf{practical controllability extension} in which click/SAM-driven target-mask acquisition replaces the original point-cloud input with a target-specific semantic-colored input, activates only the target support points, and biases selective 3D box emission on ARKitScenes.

\section{Related Work}

We review four lines of work relevant to our method: 3D reconstruction and indoor layout estimation, 2D segmentation and semantic lifting, point-cloud representations for structured scene understanding, and language-conditioned structured indoor modeling.

\subsection{3D Reconstruction and Indoor Layout Estimation}

Classical structure-from-motion and multi-view stereo pipelines reconstruct scene geometry from multiple images through feature matching, camera estimation, triangulation, and global optimization~\cite{schonberger2016structure}. Indoor layout estimation has also been studied from single images and panoramas by exploiting geometric priors such as Manhattan-world structure, wall-floor boundaries, and room-corner regularities~\cite{hedau2009recovering,lee2009geometric,zou2018layoutnet,sun2019horizonnet}. More recent structured reconstruction methods predict floorplans, wall segments, or room layouts with learned query mechanisms and graph-like representations~\cite{yue2023roomformer}. In parallel, single-view and multi-view learning-based systems recover object pose, layout, mesh geometry, or dense scene structure directly from RGB or posed image sequences~\cite{nie2020total3d,murez2020atlas,sun2021neuralrecon}.

Recent feed-forward visual geometry models further reduce the need for expensive per-scene optimization by predicting point maps, depth, camera parameters, and 3D correspondences directly from image collections~\cite{wang2024dust3r,leroy2024mast3r,wang2025vggt,wang2025pi}. Related calibration-free or uncalibrated works has also been studied through probabilistic triangulation, adaptive multi-view/temporal Transformer fusion, and CNN--Transformer fusion, which aggregate cross-view evidence without relying on explicit camera calibration~\cite{jiang2023probabilistic,shuai2023adaptive,song2026ectformer}. These models make image-to-point-cloud pipelines more practical, but they primarily focus on geometric reconstruction rather than structured symbolic output. Our work uses such reconstructed point clouds as input and focuses on improving the downstream structured prediction of walls, doors, windows, and furniture boxes through a semantic-colored point-token interface.

\subsection{2D Segmentation, Semantic Lifting, and Multimodal 3D Semantics}

2D segmentation has progressed from dense semantic prediction with fully convolutional networks~\cite{long2015fcn} to instance and panoptic segmentation frameworks based on region proposals, masks, and transformer-style mask decoding~\cite{he2017maskrcnn,cheng2021maskformer,cheng2022mask2former}. Promptable segmentation models further allow masks to be generated from points, boxes, text-like concepts, or other user-specified prompts~\cite{kirillov2023sam,carion2025sam}. These models provide strong 2D grouping cues that can complement sparse or incomplete 3D geometry.

A related line of work lifts 2D semantic, visual-language, or open-vocabulary features into 3D maps and point clouds. CLIP-style visual-language pretraining enables category-level alignment between visual regions and textual concepts~\cite{radford2021learning}, while ConceptFusion, OpenScene, OpenMask3D, LERF, and CLIP-Fields explore open-set or language-aligned 3D scene representations by associating 2D features with 3D structures~\cite{jatavallabhula2023conceptfusion,peng2023openscene,takmaz2023openmask3d,kerr2023lerf,shafiullah2023clipfields}. These works aim mainly at semantic mapping, open-vocabulary querying, or 3D instance segmentation. By contrast, our method uses 2D segmentation only as a coarse, controllable conditioning signal for structured indoor decoding. The semantic representation is intentionally reduced to furniture, walls, openings, and others, because these groups correspond directly to the layout and box errors targeted by our structured prediction task.

\subsection{Point-Cloud Representations and 3D Scene Understanding}

Point-cloud understanding has evolved from early point-wise networks to hierarchical, convolutional, and transformer-based representations. PointNet and PointNet++ introduced direct learning on unordered point sets and local hierarchical aggregation~\cite{qi2017pointnet,qi2017pointnetplusplus}. KPConv and sparse convolutional networks improved local geometric processing and scalable 3D perception~\cite{thomas2019kpconv,choy20194d}. Transformer-style point models and improved point-network training recipes further strengthened point-level recognition and segmentation~\cite{zhao2021pointtransformer,qian2022pointnext}. For 3D object detection, VoteNet and Group-Free 3D demonstrate how point features can be used to predict object centers and boxes in indoor scenes~\cite{qi2019votenet,liu2021groupfree3d}.

Self-supervised point pretraining has become increasingly important for robust 3D representation learning. Point-BERT and Point-MAE apply masked modeling ideas to point clouds~\cite{yu2022point,pang2022pointmae}, while Sonata~\citep{wu2025sonata} and Utonia~\citep{zhang2026utonia} study stronger and more general point encoders for reliable point representations across scenes and tasks. These methods provide compact and transferable 3D tokens, but they are typically optimized for recognition, segmentation, detection, or feature transfer. Our work builds on such point-token encoders, especially the SpatialLM/Sonata-style token interface, but specifically targets the under-representation of openings and cluttered furniture instances in LLM-conditioned structured indoor prediction.

\subsection{Language-Conditioned Structured Indoor Modeling}

Language models and autoregressive sequence decoders provide a flexible way to represent structured scene outputs as serialized tokens. SceneScript reconstructs scenes with an autoregressive structured language model~\cite{avetisyan2024scenescript}, and SpatialLM trains large language models for structured indoor modeling from point-cloud tokens~\cite{mao2025spatiallm}. Related 3D-language models explore how large language models can be grounded in point clouds or 3D scene representations for reasoning and interaction~\cite{hong20233dllm}. Vision-language decoders such as Qwen2-VL also demonstrate the broader potential of using language-model interfaces for structured visual reasoning~\cite{wang2024qwen2}.

Closest to our setting is SpatialLM~\cite{mao2025spatiallm}, which conditions a language model on point-cloud tokens and outputs structured indoor layouts and object lists. SpatialLM shows that many indoor modeling tasks can be formulated as language modeling over 3D tokens, but its conditioning signal remains primarily geometric. Our work follows the same decoder-side formulation while modifying the point input interface: instead of adding a separate semantic decoder or changing the output language, we inject a coarse RGBB semantic code before tokenization and apply a lightweight routed semantic shift after encoding. This preserves the original structured decoding protocol while strengthening the evidence for thin openings and cluttered furniture instances. General conditioning mechanisms such as language-guided modulation and FiLM-style feature transformations motivate the use of auxiliary signals to steer visual representations~\cite{de2017modulating,perez2018film}; however, our design differs by keeping the conditioning spatially localized at the point level and by routing the semantic residual within the point-token stream rather than using global feature modulation alone.

\begin{figure*}[!t]
    \centering
    \includegraphics[width=1\textwidth, clip, trim=0.3 5.7cm 0.3 7cm]{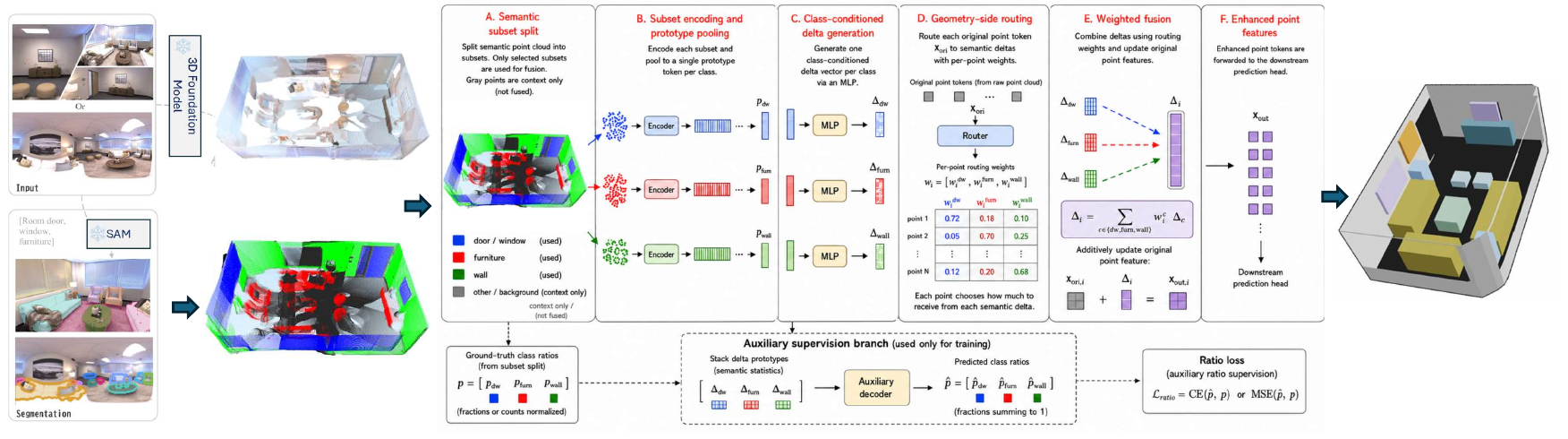}
    \caption{Overview of the proposed semantic-colored point interface. Coarse semantic cues are encoded as RGBB point attributes and injected before sparse tokenization, so geometry and semantics share the same point-token pipeline. The RGB/SAM branch illustrates how target masks can be obtained for the controllability extension, while the controlled experiments evaluate the semantic interface at the point-cloud input level. A lightweight semantic shift module injects class-conditioned residuals into the point-token stream before autoregressive structured decoding.}
    \label{fig:pipeline}
\end{figure*}

\section{Method}
\label{sec:method}

\subsection{Overview}
\label{sec:method_overview}

We follow the LLM-conditioned structured indoor prediction paradigm of SpatialLM~\cite{mao2025spatiallm}, where a language model decoder is conditioned on a sequence of 3D point tokens and autoregressively generates a serialized structured description. The central motivation is to improve fine layout elements (e.g., doors/windows) and per-instance furniture detection without changing the downstream decoder, the output serialization, or the token budget, so that gains can be attributed to the conditioning interface rather than to a new decoding protocol.

To this end, we inject coarse semantics as an explicit, localized conditioning signal on top of unchanged 3D geometry tokens (rather than as a replacement for geometry). We implement this with a tokenization-compatible semantic-colored point interface and a minimal semantic shift module that counteracts semantic attenuation introduced by voxelization and sparse pooling, while keeping the token space and decoding protocol intact. Besides improving offline accuracy, the same interface provides an interpretable control knob at inference time (Section~\ref{sec:ablation}), and serves as a sanity check against degenerate geometry-only shortcuts.

The subsequent subsections naturally follow this intent: we first define how four-group semantics are lifted and encoded into points, then describe tokenization and LLM conditioning, then introduce the lightweight shift injection and its regularization/diagnostics, and finally summarize the training objective.

\subsection{Joint Raw and RGBB Point Attributes from Semantic Sources}
\label{sec:method_prompt}
Point tokenization tends to dilute thin or boundary-dominated structures and to smear instance boundaries under sparse pooling; therefore, we want a semantic signal that is (i) easy to obtain, (ii) spatially localized, and (iii) able to survive the same voxelization/serialization path as geometry without requiring token-wise alignment after pooling. We achieve this by representing coarse semantics as RGBB point attributes and using them jointly with the original raw point input to the point tokenization backbone.

Concretely, we encode coarse semantic cues as a four-group point code (furniture / walls / openings / others) with a fixed four-color convention in RGB: R for furniture, G for walls, B for openings, and black for others. The name RGBB refers to four semantic color states encoded in a standard three-channel RGB vector; black is the all-zero RGB state, not an additional feature channel. We append this code to the original raw point attributes \emph{before} sparse tokenization, so semantics and geometry share the same serialization/pooling path and no token-wise correspondence is required after pooling.

Let $\mathcal{P}=\{(\mathbf{x}_i,\mathbf{f}_i)\}_{i=1}^{N_p}$ be the input point cloud, where $\mathbf{x}_i\in\mathbb{R}^3$ are 3D coordinates and $\mathbf{f}_i\in\mathbb{R}^{D}$ are per-point attributes.
We construct $\mathbf{f}_i$ to contain the original raw point attributes together with a \emph{semantic color code} that encodes a coarse partition with $K{=}4$ groups: furniture, walls, openings (doors/windows), and others.
We represent these groups using a four-color convention in RGB space: \textbf{R} (furniture), \textbf{G} (walls), \textbf{B} (openings), and \textbf{black} $(0,0,0)$ (others).
The semantic code is obtained from ground-truth annotations when constructing oracle RGBB inputs for controlled quantitative experiments at the point-cloud interface. For the click-based controllability demonstration at inference time, the practical branch uses a click-ready SAM~3 promptable segmentation interface~\cite{carion2025sam} to obtain a target instance mask from the reference RGB view. In our implementation, the point-cloud input is represented in the aligned point-map form used by VGGT-style point cloud maps, so each valid image pixel is associated with a 3D point. The SAM mask can therefore be applied directly to the aligned point-map pixels, and multi-view masks aggregate naturally into the four semantic groups.

Let $\mathcal{A}(i)$ denote the aligned point-map pixels associated with point $i$ across the available views.
We obtain mutually exclusive binary masks $\mathcal{M}_{v}^{\text{fur}}$, $\mathcal{M}_{v}^{\text{wall}}$, and $\mathcal{M}_{v}^{\text{op}}$ on each view $v$, aggregate category support over $(v,\mathbf{u})\in\mathcal{A}(i)$, and assign points with no foreground support to ``others''.
This defines a hard 4-way assignment $\mathbf{p}_i\in\{0,1\}^{4}$ by
\begin{equation}
\mathbf{p}_i = [p_{i,\text{fur}},p_{i,\text{wall}},p_{i,\text{op}},p_{i,\text{oth}}],
\label{eq:lift_scores}
\end{equation}
where $p_{i,k}{=}1$ for the category with the largest accumulated support among furniture, walls, openings, and others after mask exclusivity is enforced.
We then encode the semantic group as an RGB vector with a black ``others'' default:
\begin{equation}
\mathbf{c}^{\text{sem}}_i
=
[p_{i,\text{fur}},\ p_{i,\text{wall}},\ p_{i,\text{op}}]
\in\{0,1\}^{3}.
\label{eq:rgb_code}
\end{equation}
Finally, we define the point attribute as
\begin{equation}
\mathbf{f}_i = [\mathbf{g}_i;\ \mathbf{c}^{\text{sem}}_i],
\label{eq:feat_concat}
\end{equation}
where $\mathbf{g}_i$ denotes the original raw appearance together with any remaining low-dimensional attributes expected by the point backbone implementation (e.g., coordinate-related input channels).
This provides point-level semantic association while injecting semantics \emph{before} tokenization, avoiding any requirement of token-wise correspondence in latent space.

The resulting interface is deliberately minimal: it preserves the original geometric pathway and introduces a coarse, manipulable semantic cue that remains editable at the input level for controllability experiments, while keeping the downstream decoding protocol unchanged.

The above hard semantic-color construction matches the implementation used in our experiments and enables direct input manipulation by toggling semantic-colored support points.
While soft probabilities and multi-view confidence fusion are possible extensions, we do not use them in this work.

\subsection{Point Tokenization and LLM Conditioning}
\label{sec:method_token_llm}
To focus the analysis on the conditioning interface, we keep the point-token pipeline, token budget, and LLM decoding protocol unchanged while modifying the \emph{input interface} and adding only a lightweight token-level semantic shift. This design aims to attribute changes to the semantic interface rather than to a different tokenization budget, a different backbone family, or a different serialization/decoding rule.

Accordingly, we follow the point-token interface of SpatialLM~\cite{mao2025spatiallm} while changing the \emph{input point attributes} and, when enabled, applying the lightweight shift module defined below.
Concretely, we feed the point cloud $\mathcal{P}$ with joint raw and RGBB attributes into the pretrained Sonata backbone~\citep{wu2025sonata}.
Following the SpatialLM implementation, each point provides an integer grid coordinate for sparse indexing and float attributes for feature encoding.
The backbone consumes a sparse dictionary with fields \texttt{grid\_coord}, \texttt{coord}, and \texttt{feat}, and produces a token sequence
\begin{equation}
\mathbf{H} \;=\; \mathcal{E}(\mathcal{P}) \in \mathbb{R}^{N \times C},
\label{eq:token_seq}
\end{equation}
where $N$ is the number of tokens after sparse serialization/pooling and $C$ is the token dimension.
Unless otherwise stated, we use the pretrained encoder $\mathcal{E}$ to define the shared token pipeline and keep the token budget plus decoder-side protocol fixed, so performance differences are attributed to the semantic-coded input interface rather than to a different serialization strategy.

Compared to a two-stream design (separately tokenizing geometry and semantics and fusing in latent space), our construction avoids latent token alignment issues caused by voxelization/pooling: the semantic signal is injected at the point level and thus naturally follows the same sparse tokenization path as geometry.

We adopt an autoregressive language model decoder as a structured sequence generator (a Qwen2-VL model~\cite{wang2024qwen2} in our implementation) and keep the output serialization and decoding protocol identical to SpatialLM. To avoid ambiguity between the pre-shift backbone tokens $\mathbf{H}$ and the final tokens consumed by the LLM, Section~\ref{sec:method_llm} specifies the projection and autoregressive objective after the optional semantic shift is defined. Thus, this subsection defines the shared token interface and the preconditioning representation $\mathbf{H}$.

\subsection{Semantic Shift Injection Module}
\label{sec:method_shift}
Although semantic information is injected at the point level, sparse tokenization can still attenuate or smear signals due to voxel pooling and serialization, especially for openings and crowded furniture. The goal of the shift module is therefore not to redesign the backbone or introduce a separate semantic stream, but to provide a minimal adaptation layer that amplifies the injected signal in the token space while preserving the original representation and decoding protocol.

To this end, we apply a lightweight \emph{semantic shift} module on top of the point encoder output.

In our implementation, the shift module is a class-conditional residual injection with a geometry-side router.
We focus on three non-background semantic groups (furniture, walls, and openings/doors-windows), while ``others'' are encoded as black and treated as background.
Let $\mathbf{H}=[\mathbf{h}_1,\ldots,\mathbf{h}_N]^\top\in\mathbb{R}^{N\times C}$ denote geometry tokens produced by the point encoder from the raw+RGBB point cloud.
In addition, to avoid any ambiguity introduced by sparse voxel pooling, we compute three semantic prototypes via
\emph{class-filtered} forward passes through the same point encoder.
For each non-background group $k\in\{\text{furn},\text{wall},\text{op}\}$, we construct a filtered point set
$\mathcal{P}^{(k)}$ by keeping only points whose semantic color matches $k$ (dropping all others) and run the same
encoder to obtain a token set $\mathbf{H}^{(k)}$.
We then define
\begin{equation}
\mathbf{p}_{k} = \mathrm{Pool}(\mathbf{H}^{(k)}) \in \mathbb{R}^{C},\qquad k\in\{\text{furn},\text{wall},\text{op}\},
\label{eq:sem_proto}
\end{equation}
where $\mathrm{Pool}(\cdot)$ is mean pooling and we set $\mathbf{p}_k=\mathbf{0}$ if $\mathcal{P}^{(k)}$ is empty.
With $K{=}3$ foreground groups, this requires $K{+}1$ encoder forwards per scene (one full forward plus three filtered forwards),
but does not introduce any additional parameters and keeps the conditioning interface alignment-free.
For each group, we map the pooled prototype through a small MLP and define the corresponding nonnegative class delta vector as $\boldsymbol{\delta}_{k}=\phi_k(\mathbf{p}_k)\in\mathbb{R}^{C}$, where $\phi_k$ is implemented with a softplus output nonlinearity. We further restrict each $\boldsymbol{\delta}_k$ to act on a disjoint channel block to reduce interference.
For each geometry token $\mathbf{h}_i$, a router predicts routing weights $\mathbf{w}_i\in\Delta^{2}$:
\begin{equation}
\mathbf{w}_i = \mathrm{softmax}(r_{\omega}(\mathbf{h}_i))\in\mathbb{R}^{3},\qquad \sum_{k} w_{i,k}=1,
\label{eq:router}
\end{equation}
and we fuse the class deltas by a residual update
\begin{equation}
\mathbf{h}_i^{f} = \mathbf{h}_i + \alpha\sum_{k\in\{\text{furn},\text{wall},\text{op}\}} w_{i,k}\,\boldsymbol{\delta}_{k},
\label{eq:shift_fuse}
\end{equation}
where $\alpha>0$ is a fixed residual-scale hyperparameter shared across experiments.
This design preserves the original token space and avoids any token-wise correspondence between separate geometry/semantic streams, since all semantics enter through the same semantic-colored point interface \emph{before} tokenization.

Practically, the benefit is a controlled, class-aware residual adjustment that counteracts semantic attenuation without modifying the backbone or the decoder: tokens that are likely to correspond to furniture/walls/openings receive stronger, more separable conditioning, while empty-group cases are safely handled by $\mathbf{p}_k{=}\mathbf{0}$.

\paragraph{Regularizing lightweight injection}
We use three small auxiliary terms (Eq.~\eqref{eq:total}) to regularize the shift module and make its routing behavior analyzable:
(1) $\mathcal{L}_{\text{ratio}}$ encourages the predicted global semantic ratio from the auxiliary head to match the empirical ratio implied by semantic-colored points;
(2) $\mathcal{L}_{\text{budget}}$ constrains the average routing mass to stay consistent with the empirical ratio, preventing collapse to a single group;
(3) $\mathcal{L}_{\text{ent}}$ encourages non-degenerate routing by maximizing the entropy of $\mathbf{w}_i$.
Overall, these terms are used with small weights and do not change the decoding protocol. Empirically, we treat them as lightweight constraints whose main role is to reduce degenerate routing and expose layout--box trade-offs, not as a standalone stability contribution.

\subsection{Auxiliary Decoder Head for Semantic Ratio Regularization}
\label{sec:method_aux_decoder}
Because the shift module uses routing and residual injection, we need a simple mechanism to (i) monitor whether the module reacts coherently to the injected semantics and (ii) regularize against degenerate routing behaviors, while still keeping the final structured outputs produced only by the language model decoder. We therefore attach a lightweight \emph{auxiliary decoder head} inside the semantic shift module that is used for training-time regularization and diagnostics, but not for prediction.
The head produces a global distribution over the three non-background semantic groups (furniture, walls, openings); the black ``others'' group is treated as background and excluded from the ratio.
Let $\mathbf{s}\in\mathbb{R}^{6}$ denote a compact summary vector computed from class-delta statistics:
\begin{equation}
\mathbf{s} = [\mu_{\text{furn}},\mu_{\text{wall}},\mu_{\text{op}},\nu_{\text{furn}},\nu_{\text{wall}},\nu_{\text{op}}],
\label{eq:aux_stats}
\end{equation}
where $\mu_k$ and $\nu_k$ are the mean value and the $\ell_2$ norm of the corresponding channel block of $\boldsymbol{\delta}_k$ (computed per scene and concatenated).
We then predict
\begin{equation}
\hat{\mathbf{p}} = \mathrm{softmax}\big(g_{\psi}(\mathbf{s})\big) \in \Delta^{2},
\label{eq:aux_ratio_pred}
\end{equation}
where $g_{\psi}$ is a shallow MLP and $\Delta^{2}=\{\mathbf{p}\in\mathbb{R}^{3}:\; p_k\ge 0,\;\sum_{k=1}^{3}p_k=1\}$ is the probability simplex. Throughout this subsection, we use the shared foreground order $(\text{furn}, \text{wall}, \text{op})$, which matches the RGB channel order in Eq.~\eqref{eq:rgb_code}; $\hat{\mathbf{p}}$, $\mathbf{p}^{\text{gt}}$, and $\bar{\mathbf{w}}$ are all interpreted in this order.
During training, we use $\hat{\mathbf{p}}$ to define a lightweight ratio-consistency regularizer (Eq.~\eqref{eq:l_ratio}) by matching it to the empirical ratio computed from the semantic-colored points; at inference time, this head is not used to produce the final structured outputs (layout elements or 3D bounding boxes) generated by the language model decoder.
Accordingly, the performance gains we report are attributed to the point-token conditioning interface and the shift injection itself, rather than to introducing a task-specific structured prediction head.

This design keeps the overall intervention compact: it constrains the behavior of the lightweight injection during training while preserving the original end-task interface and avoiding a separate supervised prediction head at test time.

\paragraph{Inference-time controllability}
Because semantics is encoded as RGB intensities, controllability can be implemented without retraining by modifying the semantic color channels of the input point cloud.
Concretely, we consider two simple input-level controls already reflected in the Table~\ref{tab:arkit_strength_curve} controllability analysis: (i) \emph{support manipulation} (activate/deactivate semantic-colored target support points, without changing the downstream decoder), and (ii) \emph{intensity scaling} (multiply the semantic RGB by a scalar).
Both operations perturb the same conditioning interface seen during training, so they provide an interpretable knob for steering box emission frequency.
Since these controls do not alter point geometry or the decoding protocol, they also serve as a sanity check that the observed steering is not driven by a geometry-only shortcut.

\paragraph{Complexity}
The shift module adds 0.2M trainable parameters (a small router MLP, three per-class MLPs, and the auxiliary decoder head) on top of the 6.0M trainable-parameter SpatialLM baseline under our counting convention, which reports trainable parameters updated by the recipe and excludes frozen pretrained weights; its runtime and memory overhead are modest relative to the point encoder and the LLM forward pass.
This is a deliberate choice: we aim to attribute performance changes to the conditioning interface and a minimal adaptation layer rather than to a significantly larger model.

\subsection{LLM Conditioning and Training Objective}
\label{sec:method_llm}
The final component is to integrate the point tokens into an autoregressive structured generator without changing the decoder-side protocol. We therefore follow the SpatialLM conditioning interface~\cite{mao2025spatiallm} and use the auxiliary head in Section~\ref{sec:method_aux_decoder} only for training-time regularization; it does not affect the decoding protocol or the final structured outputs. When the shift module is enabled, the final token sequence is $\mathbf{H}^{f}$ from Eq.~\eqref{eq:shift_fuse}; when the shift module is disabled, we simply set $\mathbf{H}^{f}=\mathbf{H}$.
We project fused tokens $\mathbf{H}^{f}$ into the embedding space of the LLM with a linear map:
\begin{equation}
\mathbf{Z}=\mathbf{H}^{f}\mathbf{W}_{p},\qquad
\mathbf{W}_{p}\in\mathbb{R}^{C\times d},
\label{eq:proj}
\end{equation}
where $d$ is the LLM embedding dimension.
We then prepend $\mathbf{Z}$ to the textual prompt tokens $\mathbf{T}$ and train the LLM to autoregressively generate the serialized target sequence $\mathbf{Y}$:
\begin{equation}
\mathcal{L}_{\text{AR}}
=
-\sum_{t}\log p_{\Theta}\!\left(y_t \mid y_{<t},\,\mathbf{Z},\,\mathbf{T}\right),
\label{eq:ar}
\end{equation}
where $\Theta$ includes the learnable projection, the semantic shift injection module (Section~\ref{sec:method_shift}), and (when enabled) the LLM parameters.

The overall training objective in our implementation is
\begin{equation}
\mathcal{L}
=
\mathcal{L}_{\text{AR}}
+
\lambda_{\text{ratio}}\mathcal{L}_{\text{ratio}}
+
\lambda_{\text{budget}}\mathcal{L}_{\text{budget}}
-
\lambda_{\text{ent}}\mathcal{L}_{\text{ent}},
\label{eq:total}
\end{equation}
with small weights $\lambda_{\cdot}$ (fixed across all runs unless specified).

Overall, this objective provides a clear separation of roles: $\mathcal{L}_{\text{AR}}$ trains the structured generator under the unchanged serialization protocol, while the auxiliary terms act as lightweight routing constraints that help us analyze how semantic injection affects the layout--box trade-off without introducing new inference-time branches.

\paragraph{Explicit form of regularizers}
To avoid under-specifying the additional terms in Eq.~\eqref{eq:total}, we make their implementation explicit.
Let $\mathbf{w}_i\in\mathbb{R}^{3}$ be the router weights in Eq.~\eqref{eq:router} and let $\bar{\mathbf{w}}=\frac{1}{N}\sum_{i=1}^{N}\mathbf{w}_i$ be the mean routing mass.
We compute an empirical semantic ratio $\mathbf{p}^{\text{gt}}\in\Delta^{2}$ from the RGBB semantic point assignments by counting points in the three non-background groups (furniture, walls, openings) and normalizing in that same order.
We then define
\begin{equation}
\mathcal{L}_{\text{ratio}} = D_{\mathrm{KL}}\!\left(\mathbf{p}^{\text{gt}}\;\|\;\hat{\mathbf{p}}\right),
\label{eq:l_ratio}
\end{equation}
\begin{equation}
\mathcal{L}_{\text{budget}} = \left\|\bar{\mathbf{w}}-\mathbf{p}^{\text{gt}}\right\|_2^2,
\label{eq:l_budget}
\end{equation}
and an entropy-style term that we \emph{maximize} to discourage routing collapse:
\begin{equation}
\mathcal{L}_{\text{ent}} = \frac{1}{N}\sum_{i=1}^{N}\left(-\sum_{k=1}^{3} w_{i,k}\log(w_{i,k}+\epsilon)\right).
\label{eq:l_ent}
\end{equation}
These definitions match the qualitative roles described in Section~\ref{sec:method_shift}: $\mathcal{L}_{\text{ratio}}$ aligns predicted and empirical semantic ratios, $\mathcal{L}_{\text{budget}}$ prevents degenerate global routing, and $\mathcal{L}_{\text{ent}}$ encourages non-collapsed token-wise routing. In all experiments, $\epsilon$ denotes a fixed numerical stabilizer and $\alpha$ is fixed once as a shared residual-scale hyperparameter rather than being retuned per benchmark.

We initialize from pretrained point backbones throughout.
Unless explicitly noted, the experiments keep the token budget, decoding protocol, and output serialization fixed; whether a pretrained backbone is kept fixed or jointly optimized follows the corresponding training recipe, and ablations remove individual semantic-interface components while matching the remaining setup.


\section{Experiments}
\label{sec:exp}

This section evaluates the proposed semantic-colored interface as a controlled point-cloud-input semantic-conditioning mechanism for LLM-conditioned structured indoor prediction.
We study \textbf{Structured3D}~\cite{zheng2020structured3d} and the \textbf{SpatialLM dataset}~\cite{mao2025spatiallm} for \emph{point-cloud-to-structure} decoding, and \textbf{ARKitScenes}~\cite{baruch2021arkitscenes} for point-cloud-to-3D-box decoding from fixed point inputs. The matched semantic-input controls instantiate the pipeline at the point-cloud interface, while the click/SAM study evaluates the practical RGB-view mask-acquisition branch.
We compare against the LLM-based baseline \textbf{SpatialLM}~\cite{mao2025spatiallm} and the structured generation baseline \textbf{SceneScript}~\cite{avetisyan2024scenescript}, then expand the evaluation with controlled interface tests, backbone-transfer studies, efficiency accounting, inference-time controllability, and targeted failure-mode slices to characterize the mechanism's gains, boundaries, and failure modes.

\subsection{Experimental Setup}
\label{sec:exp_setup}

\paragraph{Benchmarks and evaluation regimes}
\textbf{Structured3D}~\cite{zheng2020structured3d} evaluates \emph{layout element} prediction (walls/doors/windows) in a point-cloud-to-structure setting.
\textbf{SpatialLM dataset}~\cite{mao2025spatiallm} is the canonical benchmark for LLM-conditioned structured indoor modeling. In Table~\ref{tab:main_compare_standard}, we separate its layout and furniture-box metrics for readability; in Tables~\ref{tab:main_compare_controlled}, \ref{tab:ablation_main}, and~\ref{tab:controlled_ablations}, we additionally report the official \emph{full-house} score, which jointly reflects layout and furniture-box quality under the original serialization and matching protocol.
\textbf{ARKitScenes}~\cite{baruch2021arkitscenes} provides real-world oriented 3D furniture boxes; in our main comparison, it is evaluated as point-cloud-to-box prediction from fixed point inputs.
We report two complementary comparison tables. Table~\ref{tab:main_compare_standard} provides a broad three-way summary across SceneScript, SpatialLM, and our semantic-interface variant under the semantic-source setting used in each benchmark block; it should therefore be read as a condition-specific summary rather than as a fully raw-input-only comparison. Table~\ref{tab:main_compare_controlled} reports the matched semantic-source controls: raw+RGBB controlled point-cloud results on Structured3D, the SpatialLM dataset, and ARKitScenes.

\paragraph{Inputs, semantic sources, and backbone setup}
On Structured3D and the SpatialLM dataset, the model takes a point cloud and decodes layout elements plus furniture boxes.
On ARKitScenes, the same structured decoder predicts oriented 3D furniture boxes from fixed point-cloud inputs.
Unless otherwise stated, we append a four-group semantic color code (\textbf{R}/\textbf{G}/\textbf{B}/black for furniture/walls/openings/others) to the original raw point attributes.
This semantic cue is injected \emph{before tokenization}, so semantics and geometry share the same sparse serialization/pooling path and no token-wise latent alignment is required after pooling.
For controlled quantitative results, the semantic colors come from dataset annotations and are evaluated at the point-cloud interface. These oracle or annotation-derived semantic-source settings are used to isolate the conditioning mechanism and are reported separately from the practical click/SAM branch. The click/SAM study uses a reference-view target mask from the click-ready SAM~3 interface and applies it to the aligned point-map representation to construct a target-specific semantic-colored input.
We tokenize points with the pretrained Sonata backbone~\citep{wu2025sonata} used by SpatialLM~\cite{mao2025spatiallm}. Unless explicitly stated, comparisons use the same token budget, decoding protocol, and output format so that changes reflect the semantic interface rather than decoder-side modifications.

\paragraph{Training protocol and compute budget}
We follow the SpatialLM conditioning interface~\cite{mao2025spatiallm} to project point tokens into the LLM embedding space and train the autoregressive structured decoder.
The point backbone is initialized from pretrained weights. Depending on the recipe, it may be jointly optimized together with the LLM-side projection/decoder and the lightweight semantic modules; throughout, we keep the decoding and output format unchanged across methods.
Unless explicitly noted, parameter counts in this paper report the complete set of trainable parameters under the corresponding recipe, i.e., every parameter updated by the optimizer in that run. If a pretrained backbone or decoder block is frozen by the recipe, it is not counted as trainable; if it is updated, it is included.
Unless otherwise stated, runs use \textbf{4$\times$ NVIDIA RTX PRO 6000 GPUs} with mixed precision and distributed data parallelism, and Table~\ref{tab:efficiency_overhead} reports the concrete wall-clock times for each backbone/variant.
We additionally report a reference baseline trained with the original \textbf{SpatialLM} recipe to separate interface effects from recipe-driven gains.

\subsection{Evaluation Metrics}
\label{sec:metrics}

\paragraph{Layout metrics}
For layout elements (walls, doors, windows), we report F1 at IoU$_{2D}$ thresholds 0.25 and 0.5 using the SpatialLM matching protocol~\cite{mao2025spatiallm}.
A predicted element is counted as correct when its best-matched ground-truth element exceeds the corresponding IoU threshold.

\paragraph{Furniture-box metrics}
For furniture boxes, we report F1 at IoU$_{3D}$ thresholds 0.25 and 0.5 following the ARKitScenes protocol~\cite{baruch2021arkitscenes}.
A prediction is matched to the best ground-truth box of the same category and counted as correct only if the IoU$_{3D}$ threshold is satisfied.
We additionally report a \emph{miss rate} at the @0.25 operating point to make missed predictions explicit. For Structured3D, Miss@0.25 is computed from unmatched layout elements under IoU$_{2D}$@0.25; for the SpatialLM dataset, it is the complementary miss rate under the official full-house criterion at @0.25; and for ARKitScenes, it is the miss rate under IoU$_{3D}$@0.25. Table~\ref{tab:main_compare_standard} reports only the primary accuracy metrics, while Table~\ref{tab:main_compare_controlled} adds the miss-rate column on the controlled ARKitScenes block.

\paragraph{SpatialLM full-house score}
For the SpatialLM dataset, we additionally report the official full-house aggregate under the same decoding and serialization rules~\cite{mao2025spatiallm}; it is shown as Thr@0.25 and Thr@0.5 in Tables~\ref{tab:main_compare_controlled},~\ref{tab:ablation_main}, and~\ref{tab:controlled_ablations}. We use two raw references for different purposes. The ``SpatialLM (raw input, original finetuned log)'' row in Table~\ref{tab:main_compare_controlled} retains the finetuned raw-input evaluation log from the original pipeline for context and therefore reports that run only in official full-house form rather than as separate layout/object blocks. By contrast, Table~\ref{tab:backbone_swap} uses recipe-matched raw baselines that are retrained under the same protocol as the corresponding backbone variant, so comparisons in that table should be made only within each backbone block. Table~\ref{tab:main_compare_standard} instead separates SpatialLM layout and box metrics.

\subsection{Main Comparisons}
\label{sec:exp_main}
\begin{figure*}[t]
    \centering
    \includegraphics[width=\textwidth, clip, trim=0cm 1.6cm 0cm 2.5cm]{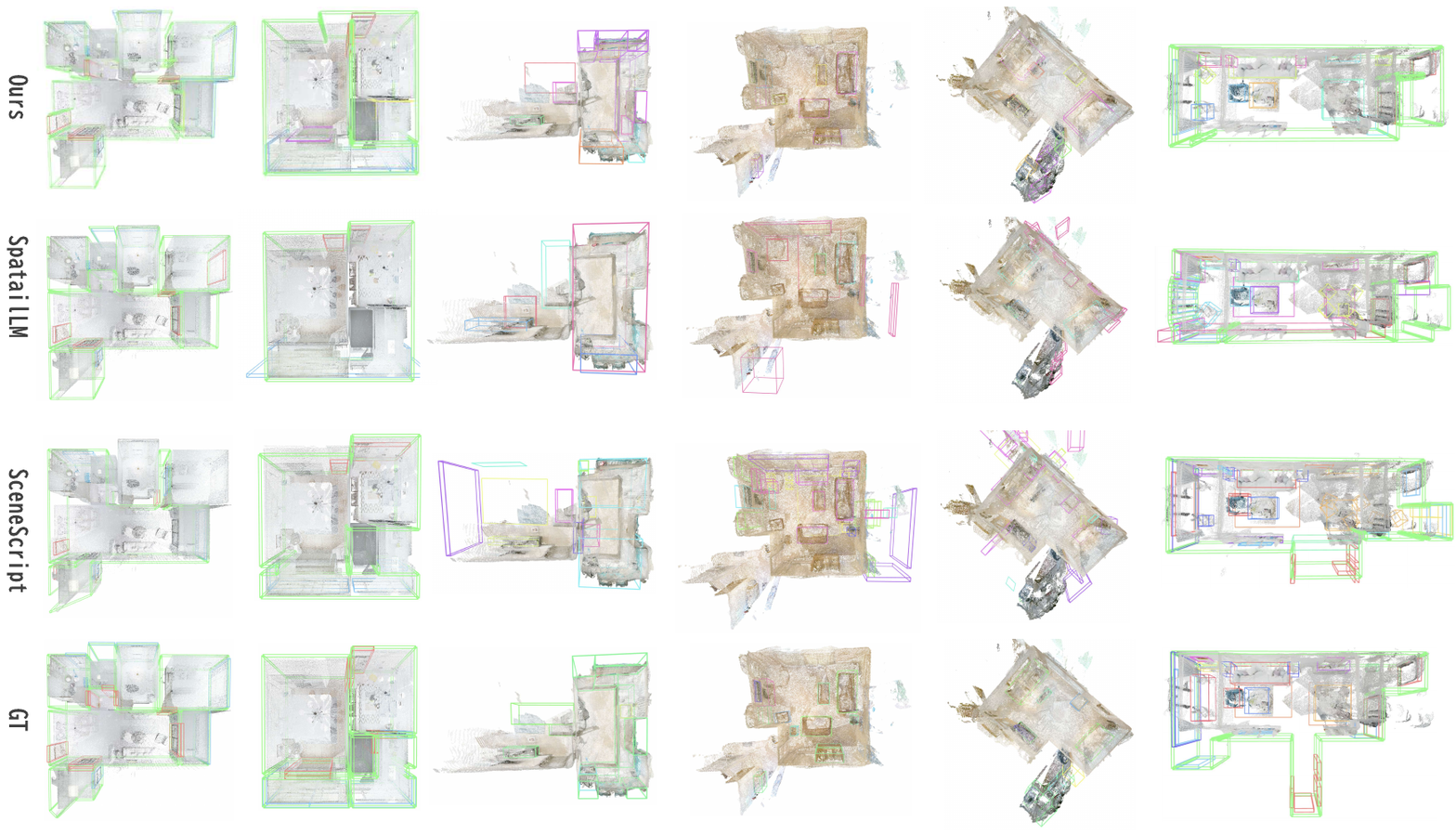}
    \caption{Qualitative comparison on three benchmarks. On Structured3D and the SpatialLM dataset, we visualize point-cloud-to-structure predictions (layout elements and 3D furniture boxes). On ARKitScenes, we visualize point-cloud-to-bounding-box results under the matched semantic point-input setting. The semantic-colored point interface (R/G/B/black for furniture/walls/openings/others) improves opening localization and reduces missed furniture detections, particularly in cluttered multi-object scenes.}
    \label{fig:comparison}
\end{figure*}

We begin with three complementary comparison tables.
Table~\ref{tab:main_compare_standard} is the broad cross-method summary under the semantic-source settings used in each benchmark block.
Table~\ref{tab:main_compare_controlled} provides the controlled counterpart with matched semantic-source settings for Structured3D, the SpatialLM dataset, and ARKitScenes.
Table~\ref{tab:backbone_swap} tests backbone transfer by replacing Sonata~\citep{wu2025sonata} with the newer Utonia encoder~\citep{zhang2026utonia}.
Figure~\ref{fig:comparison} provides representative qualitative examples from all three benchmarks.

\begin{table*}[t]
\centering
\caption{\textbf{Broad three-way comparison with SceneScript, SpatialLM, and ours under the semantic-source settings used in each benchmark block.} These controlled comparisons instantiate the pipeline at the point-cloud interface: Structured3D and the SpatialLM dataset report point-cloud-to-structure results, while ARKitScenes reports point-cloud-to-box prediction from fixed point inputs. For the SpatialLM dataset, we separate layout and box metrics instead of reporting only the full-house aggregate. The \textbf{Ours} row is condition-specific: it uses annotation-derived RGBB on Structured3D / SpatialLM and oracle raw+RGBB point inputs on ARKitScenes, so the table isolates the value of the semantic interface rather than a fully practical RGB/SAM acquisition pipeline. The parameter column reports trainable parameters under the corresponding recipe. All metrics are higher-is-better.}
\label{tab:main_compare_standard}
\tablefontsize
\setlength{\tabcolsep}{3pt}
\renewcommand{\arraystretch}{1.10}
\begin{tabular}{l|c|cc|cccc|cc}
\hline
\multirow{3}{*}{\textbf{Method}} & \multirow{3}{*}{\textbf{\#Trainable Params}} &
\multicolumn{2}{c|}{\textbf{Structured3D}} &
\multicolumn{4}{c|}{\textbf{SpatialLM dataset}} &
\multicolumn{2}{c}{\textbf{ARKitScenes}} \\
& & \multicolumn{2}{c|}{\textbf{Layout}} & \multicolumn{2}{c|}{\textbf{Layout}} & \multicolumn{2}{c|}{\textbf{BBox}} & \multicolumn{2}{c}{\textbf{Box}} \\
& & \textbf{F1@0.25} & \textbf{F1@0.5} & \textbf{F1@0.25} & \textbf{F1@0.5} & \textbf{F1@0.25} & \textbf{F1@0.5} & \textbf{F1@0.25} & \textbf{F1@0.5} \\
\hline
SceneScript & 5.4M & 0.3572 & 0.2368 & 0.5330 & 0.4418 & 0.3042 & 0.1523 & 0.3042 & 0.1922 \\
SpatialLM & 6.0M & 0.5919 & 0.4982 & 0.5644 & 0.4972 & 0.3660 & 0.2104 & 0.3662 & 0.2306 \\
Ours & 6.2M & \textbf{0.6340} & \textbf{0.5986} & \textbf{0.5954} & \textbf{0.5410} & \textbf{0.4300} & \textbf{0.2907} & \textbf{0.4013} & \textbf{0.3417} \\
\hline
\end{tabular}
\end{table*}

\begin{table*}[t]
\centering
\caption{\textbf{Controlled comparisons in the three-benchmark format.} Structured3D and SpatialLM dataset columns report point-cloud-to-structure results under the raw+RGBB oracle interface, while ARKitScenes reports point-cloud-to-box results under the matched oracle-RGBB point-input control. The SpatialLM raw-input reference row is retained for context. ``--'' denotes an omitted benchmark/variant pair. Miss@0.25 is lower-is-better; all other metrics are higher-is-better.}
\label{tab:main_compare_controlled}
\tablefontsize
\setlength{\tabcolsep}{4pt}
\renewcommand{\arraystretch}{1.10}
\begin{tabular}{l|cc|cc|ccc}
\hline
\multirow{2}{*}{\textbf{Method}} &
\multicolumn{2}{c|}{\textbf{Structured3D: Layout}} &
\multicolumn{2}{c|}{\textbf{SpatialLM dataset: Full-house}} &
\multicolumn{3}{c}{\textbf{ARKitScenes: Box}} \\
& \textbf{F1@0.25} & \textbf{F1@0.5} & \textbf{Thr@0.25} & \textbf{Thr@0.5} & \textbf{F1@0.25} & \textbf{F1@0.5} & \textbf{Miss@0.25$\downarrow$} \\
\hline
SpatialLM (raw input, original finetuned log) & 0.6213 & 0.5315 & 0.3348 & 0.2871 & 0.2997 & 0.2053 & 0.7280 \\
SpatialLM (finetuned+RGBB (GT)) & 0.6234 & 0.5896 & 0.3947 & 0.3201 & 0.4021 & 0.3267 & 0.6456 \\
Ours (oracle raw+RGBB + shift) & 0.6340 & 0.5986 & 0.4515 & 0.3236 & 0.4013 & 0.3417 & 0.6438 \\
\hline
\end{tabular}
\end{table*}

\paragraph{Backbone transfer as a main-table supplement}
Table~\ref{tab:backbone_swap} complements the semantic-source controls by testing whether the gains persist when replacing Sonata~\citep{wu2025sonata} with Utonia~\citep{zhang2026utonia} under matched raw-vs-full settings.

\begin{table*}[!t]
\centering
\caption{\textbf{Backbone transfer supplement to the main comparison.} Replace Sonata~\citep{wu2025sonata} with the newer Utonia backbone~\citep{zhang2026utonia} and report the matched raw baseline versus raw + RGBB + shift. Entries in parentheses report the absolute gain of the full model over the raw baseline under the same backbone. To keep the table compact, we report the @0.25 operating point on each benchmark.}
\label{tab:backbone_swap}
\tablefontsize
\setlength{\tabcolsep}{5pt}
\renewcommand{\arraystretch}{1.10}
\begin{tabularx}{\textwidth}{l|Y|c|c|c}
\hline
\textbf{Backbone} & \textbf{Variant} & \textbf{Structured3D Layout F1@0.25} & \textbf{SpatialLM Thr@0.25} & \textbf{ARKitScenes Box F1@0.25} \\
\hline
Sonata & baseline & 0.5919 & 0.4108 & 0.3662 \\
Sonata & RGBB + shift (full) & 0.6340 (+0.0421) & 0.4515 (+0.0407) & 0.4013 (+0.0351) \\
\hline
Utonia & baseline & 0.5739 & 0.3970 & 0.3970 \\
Utonia & RGBB + shift (full) & 0.5862 (+0.0123) & 0.4175 (+0.0205) & 0.4127 (+0.0157) \\
\hline
\end{tabularx}
\end{table*}

Table~\ref{tab:main_compare_standard} is the primary cross-method comparison under the semantic-source settings used in each benchmark block. Structured3D and the SpatialLM dataset are reported as point-cloud-to-structure results, with the SpatialLM dataset separated into layout and box metrics; ARKitScenes is reported as point-cloud-to-box prediction under the matched semantic point-input setting. The \textbf{Ours} row is condition-specific: on Structured3D / SpatialLM it uses annotation-derived RGBB, and on ARKitScenes it reports the matched oracle-RGBB point-input control.
Numerically, Table~\ref{tab:main_compare_standard} shows improvements over SpatialLM on the reported Structured3D, SpatialLM dataset, and ARKitScenes metrics under these condition-specific semantic-source settings. The controlled comparisons and diagnostics below then show how the gains depend on semantic source, interface variant, and benchmark setting, rather than claiming that a fully practical RGB/SAM pipeline uniformly dominates the raw baselines.
Table~\ref{tab:main_compare_controlled} keeps the same three-benchmark reading order while reporting matched point-input controls; the RGB-view click/SAM practical extension is isolated to Table~\ref{tab:arkit_strength_curve}. Table~\ref{tab:backbone_swap} shows that the full interface improves over the matched raw baseline under both Sonata and Utonia, with gains of +0.0421/+0.0407/+0.0351 on Sonata and +0.0123/+0.0205/+0.0157 on Utonia across Structured3D / SpatialLM / ARKitScenes at the @0.25 operating point.
Figure~\ref{fig:comparison} qualitatively matches this picture: openings are more precisely localized and cluttered furniture scenes show fewer missed or hallucinated boxes.
The next subsection unpacks these gains with component removals, controlled interface variants, efficiency accounting, and no-retraining controllability / failure-mode tests.

\subsection{Ablations and Analysis}
\label{sec:ablation}
We next unpack the gains along five axes: component removal, controlled interface diagnostics, efficiency overhead, inference-time controllability, and failure-mode slices.
All training-based variants keep the same data split, decoding protocol, and evaluation pipeline.
Analyses marked as no-retraining leave the trained model fixed and manipulate only the semantic support presented at inference time.

\paragraph{Component ablations under oracle raw+RGBB input}
Table~\ref{tab:ablation_main} probes the role of the train-time additions on top of the shared raw+RGBB interface: the ratio, budget, and entropy constraints together with the shift module. To make comparison easier, we present these ablations in the same three-dataset layout used above; ``--'' marks omitted benchmark/variant pairs.
The color-perturbation row is different: it is an inference-only stress test that directly perturbs the semantic code while keeping the trained model fixed.

\begin{table*}[t]
\centering
\caption{Unified component ablations in the three-dataset format. ``--'' marks omitted benchmark/variant pairs. $\dagger$: no retraining (inference-time input edits); others are trained with identical budget and decoding/output protocol.}
\label{tab:ablation_main}
\tablefontsize
\setlength{\tabcolsep}{6pt}
\renewcommand{\arraystretch}{1.10}
\begin{tabular}{l|cc|cc|cc}
\hline
\multirow{2}{*}{\textbf{Variant}} &
\multicolumn{2}{c|}{\textbf{Structured3D: Layout}} &
\multicolumn{2}{c|}{\textbf{SpatialLM dataset: Full-house}} &
\multicolumn{2}{c}{\textbf{ARKitScenes: Box}} \\
 & \textbf{F1@0.25} & \textbf{F1@0.5} & \textbf{Thr@0.25} & \textbf{Thr@0.5} & \textbf{F1@0.25} & \textbf{F1@0.5} \\
\hline
Ours (full) & 0.6340 & 0.5986 & 0.4515 & 0.3236 & 0.4013 & 0.3417 \\
\hline
w/o ratio head (disable $\mathcal{L}_{\text{ratio}}$) & 0.6034 & 0.5582 & 0.4001 & 0.3008 & 0.1482 & 0.1086 \\
w/o budget (disable $\mathcal{L}_{\text{budget}}$) & 0.6072 & 0.5046 & 0.3695 & 0.2311 & 0.2216 & 0.1408 \\
w/o entropy (disable $\mathcal{L}_{\text{ent}}$) & 0.6478 & 0.5851 & 0.4496 & 0.3083 & 0.3338 & 0.2608 \\
\hline
color perturbation on semantic RGB $\dagger$ & 0.5631 & 0.4372 & 0.3381 & 0.2819 & 0.2960 & 0.2370 \\
\hline
\end{tabular}
\end{table*}

The table supports two takeaways.
First, within the component-ablation group, the full model gives the strongest ARKitScenes box accuracy and remains competitive on Structured3D layout; the no-entropy variant improves Structured3D layout scores but sacrifices ARKitScenes box accuracy, indicating a layout--box trade-off rather than a universally superior ``stability'' objective.
Second, the inference-time semantic-color perturbation stress test also hurts both Structured3D layout and ARKitScenes box prediction, confirming that the trained decoder is genuinely using the semantic signal rather than ignoring it.

\paragraph{Controlled interface diagnostics}
Table~\ref{tab:controlled_ablations} keeps reconstruction output, token budget, and decoding/output protocol fixed while changing only the semantic interface or the presence of the shift module.
These variants diagnose semantic source, color coding, and fusion strategy; they are not designed to eliminate every color-related confound.
Three comparisons are the most interpretable. First, the random four-color control is weaker than the semantic RGBB variant in every reported column (for example, Structured3D layout F1@0.25: 0.5950 vs. 0.6340; SpatialLM Thr@0.25: 0.4153 vs. 0.4515; ARKitScenes box F1@0.25: 0.3701 vs. 0.4013), but it remains a useful control because it tests whether channelized color itself explains the gains. The consistent gap to semantic RGBB shows that the category assignment provides information beyond the color-channel prior. Second, naive post-encoder fusion is consistently weaker than the shared point-level interface, especially on ARKitScenes, which is consistent with latent token misalignment after sparse pooling. Third, comparing the full model with the same raw+RGBB input without the shift module shows that ARKitScenes F1@0.25 is nearly tied, while F1@0.5 favors the full model; the clearest gain of the routed shift appears on the SpatialLM full-house metrics.

\begin{table*}[t]
\centering
\caption{Controlled interface diagnostics. The token budget and decoding/output protocol are fixed; only the semantic interface, token-fusion strategy, or shift components vary. Raw-only baselines are reported separately in Tables~\ref{tab:main_compare_standard} and~\ref{tab:main_compare_controlled} so that every row here remains a matched interface variant.}
\label{tab:controlled_ablations}
\tablefontsize
\setlength{\tabcolsep}{4pt}
\renewcommand{\arraystretch}{1.10}
\begin{tabularx}{\textwidth}{Y|cc|cc|cc}
\hline
\multirow{2}{*}{\textbf{Variant}} &
\multicolumn{2}{c|}{\textbf{Structured3D: Layout}} &
\multicolumn{2}{c|}{\textbf{SpatialLM dataset: Full-house}} &
\multicolumn{2}{c}{\textbf{ARKitScenes: Box}} \\
& F1@0.25 & F1@0.5 & Thr@0.25 & Thr@0.5 & F1@0.25 & F1@0.5 \\
\hline
Ours (raw + RGBB + shift) & 0.6340 & 0.5986 & 0.4515 & 0.3236 & 0.4013 & 0.3417 \\
Random 4-color code & 0.5950 & 0.5622 & 0.4153 & 0.3076 & 0.3701 & 0.3121 \\
Naive post-encoder fusion (raw token + semantic token) & 0.5842 & 0.4766 & 0.4496 & 0.3083 & 0.3086 & 0.2627 \\
Same raw+RGBB input w/o shift module & 0.6234 & 0.5896 & 0.3947 & 0.3201 & 0.4021 & 0.3267 \\
\hline
\end{tabularx}
\end{table*}

Taken together, these tests provide a diagnostic map of the interface: they show when semantic coloring helps, when sharing the point-level tokenization path matters, when the routed shift adds value beyond the same raw+RGBB input, and where color/entropy confounds remain intertwined with semantic effects.

\paragraph{Efficiency overhead}
Accuracy gains are useful only if the added semantic path remains lightweight. Table~\ref{tab:efficiency_overhead} therefore reports training time, inference latency, peak GPU memory, and the encoder-forward overhead induced by evaluating the original geometry stream together with additional semantic supports.

\begin{table}[!htbp]
\centering
\caption{Efficiency overhead and encoder-forward cost. $K$ denotes the number of additional semantic-support forwards; the full semantic interface therefore incurs $K{+}1$ encoder forwards per scene.}
\label{tab:efficiency_overhead}
\tablefontsize
\setlength{\tabcolsep}{3pt}
\renewcommand{\arraystretch}{1.05}
\begin{tabular}{p{0.33\columnwidth}|cccc}
\hline
\textbf{Variant} & \textbf{Train h} & \textbf{Lat. s} & \textbf{Mem. GB} & \textbf{Fwds} \\
\hline
Sonata raw baseline & 8.64 & 0.151 & 33 & 1 \\
Sonata raw + RGBB + shift & 9.49 & 0.171 & 39 & $K{+}1$ \\
Utonia raw baseline & 13.90 & 0.180 & 36 & 1 \\
Utonia raw + RGBB + shift & 14.62 & 0.182 & 42 & $K{+}1$ \\
\hline
\end{tabular}
\end{table}

\paragraph{Inference-time controllability on ARKitScenes}
\begin{figure}[!t]
\centerline{\includegraphics[width=0.5\textwidth, clip, trim=2cm 3cm 2cm 5cm]{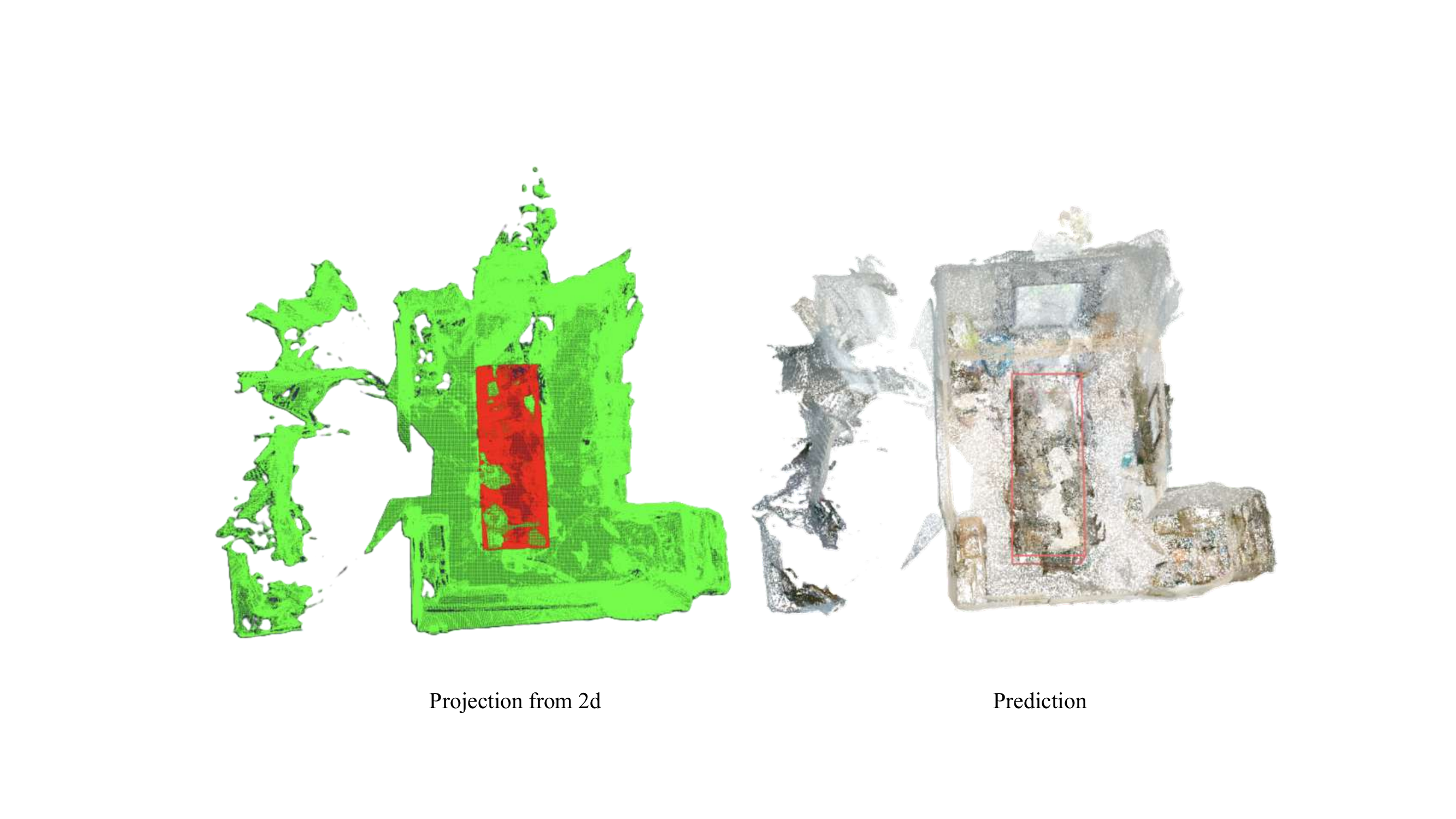}}
\caption{Controllability via click-based 2D segmentation. A user clicks on a target object to obtain a 2D instance mask via the SAM interface (left). The mask is associated with the aligned point-cloud representation to form a target-specific semantic-colored input (middle), where target support points carry the active semantic color and all non-target points use the black semantic state. The fixed pipeline then produces the corresponding oriented 3D bounding box (right). This branch is evaluated as the click/SAM controllability setting.}
\label{fig:abl2}
\end{figure}

We report two no-retraining ARKitScenes analyses because they directly test whether the semantic-colored interface can serve as a useful, but imperfect, control knob for the practical mask-acquisition branch.
Figure~\ref{fig:abl2} illustrates the intervention: the already trained pipeline receives a target-specific semantic-colored point input, while downstream reconstruction, decoding, serialization, and parsing components are kept fixed. In our aligned point-map representation, the original point input and the semantic-colored point input carry the same geometric support; the intervention therefore changes the target-specific semantic support presented to the fixed decoder.
First, in a target-only study, we randomly choose one visible furniture bounding box per test scene, yielding one target per scene and 42 clicked targets in total. We then use a click-ready SAM~3 promptable segmentation interface to obtain the selected target instance mask from the reference RGB view. The mask is associated with the aligned point-cloud representation to identify the target support points.
We construct the target-specific semantic input so that \emph{only} the aligned target support points carry an active semantic color and \emph{all other points use the black semantic state}.
We evaluate (i) \emph{target hit-rate}, i.e., whether the decoded output contains at least one predicted furniture box matched to the target ground-truth instance at IoU$_{3D}$@0.25, and (ii) \emph{target-only success}, i.e., whether the target is hit and the decoded output contains \emph{no additional} predicted furniture boxes.
Concretely, we parse the decoded sequence into a set of predicted furniture boxes using the same SpatialLM serialization/parsing pipeline, and define target-only success as: \#(predicted furniture boxes) $=1$ and that single box matches the target at IoU$_{3D}$@0.25.
Here \#(predicted furniture boxes) counts \emph{all decoded boxes}, including unmatched false positives; thus, emitting any extra box (matched or not) breaks target-only success.
Second, we report a \emph{control-strength curve} by scaling semantic color intensity to 100\%/75\%/50\%/25\% at inference time on this same clicked-target set, reusing the same selected targets and aligned masks while varying only semantic RGB intensity. For the F1 curve, we also include a point-input reference evaluated on the same clicked-target set without click/SAM target-specific semantic control. For the controllability metrics, all settings use the 100\% target-specific semantic input as the reference point. Specifically, $\Delta$FP denotes the increase in the average number of non-target false-positive boxes per scene relative to that 100\% reference, and \emph{precision drop} denotes the absolute decrease in box precision at IoU$_{3D}$@0.25 relative to the same reference.

\begin{table}[!htbp]
\centering
\caption{ARKitScenes control-strength curve. Scale semantic-color intensity (100/75/50/25\%) at inference and report box F1 on the same clicked-target set used for the target-only study. The clicked-target set contains 42 randomly selected furniture targets, one per test scene. The point-input reference uses the same clicked-target set without click/SAM target-specific semantic control.}
\label{tab:arkit_strength_curve}
\tablefontsize
\setlength{\tabcolsep}{3pt}
\renewcommand{\arraystretch}{1.12}
\begin{tabular}{lccccc}
\hline
\textbf{Input} & \shortstack{\textbf{Point-input}\\\textbf{reference}} & \textbf{100\%} & \textbf{75\%} & \textbf{50\%} & \textbf{25\%} \\
\hline
F1@IoU$_{3D}$@0.25 & 0.8041 & 0.8421 & 0.6467 & 0.3452 & 0.2105 \\
F1@IoU$_{3D}$@0.5  & 0.6940 & 0.7158 & 0.5200 & 0.1929 & 0.0000 \\
\hline
\end{tabular}
\end{table}

Table~\ref{tab:controllability_metrics} quantifies the controllability claim by reporting target hit-rate, target-only success, $\Delta$FP, and precision drop across the same inference-time intensity sweep.

\begin{table}[!htbp]
\centering
\caption{Controllability metrics under semantic-intensity scaling on the same clicked-target set of 42 randomly selected furniture targets, one per test scene. $\Delta$FP is the increase in the average number of non-target false-positive boxes per scene relative to the 100\% setting, and precision drop is the absolute decrease in precision at IoU$_{3D}$@0.25 relative to the same reference. Higher target hit-rate / target-only success are better; lower $\Delta$FP and precision drop are better.}
\label{tab:controllability_metrics}
\tablefontsize
\setlength{\tabcolsep}{4pt}
\renewcommand{\arraystretch}{1.10}
\begin{tabular}{lcccc}
\hline
\textbf{Metric} & \textbf{100\%} & \textbf{75\%} & \textbf{50\%} & \textbf{25\%} \\
\hline
Target hit-rate & 0.9524 & 0.8810 & 0.8095 & 0.2381 \\
Target-only success & 0.8810 & 0.7143 & 0.5952 & 0.1667 \\
$\Delta$FP & 0.0000 & +1.9524 & +2.5714 & +0.7143 \\
Precision drop & 0.0000 & +0.4113 & +0.5354 & +0.5660 \\
\hline
\end{tabular}
\end{table}

Together with Tables~\ref{tab:arkit_strength_curve} and~\ref{tab:controllability_metrics}, these analyses ask whether manipulating semantic support merely increases box count or instead steers predictions in a targeted manner, with the point-input reference separating the clicked-target subset from the target-specific semantic intervention.
The clearest positive signal comes from the 100\% target-only support setting. By contrast, the intensity-scaling sweep is a stress test of control strength, because lower intensities also incur substantial non-target false positives and precision loss.

\paragraph{Failure-mode slices}
Finally, Table~\ref{tab:failure_mode_slices} slices evaluation by three failure modes directly targeted by the proposed interface: thin openings, cluttered furniture, and low-support scenes (point sparsity / view coverage). We construct these as diagnostic, non-mutually-exclusive slices using dataset-derived scene statistics computed only on the evaluation split. Operationally, thin openings use scenes in the bottom quartile of annotated door/window span, cluttered furniture uses scenes in the top quartile of furniture-instance count, and low-support scenes use scenes in the bottom quartile of visible point count / view coverage after reconstruction. When ties occur at a quartile boundary, all tied scenes are included, so slice sizes can differ slightly across benchmarks. These slices separate average-case gains from the targeted robustness claim of the paper. Sensitivity to click-mask quality and lifting noise is analyzed through the no-retraining controllability results above.

\begin{table}[!htbp]
\centering
\caption{Failure-mode slices targeted by the proposed semantic interface. Slices are constructed independently (i.e., they are not mutually exclusive) from opening-width, furniture-density, and support/coverage statistics within the evaluation split. Bottom-quartile opening width defines thin openings, top-quartile furniture-instance count defines cluttered furniture, and bottom-quartile visible point count / view coverage defines low support; boundary ties are included. Each row uses an independently reported primary metric aligned with that slice.}
\label{tab:failure_mode_slices}
\tablefontsize
\setlength{\tabcolsep}{3pt}
\renewcommand{\arraystretch}{1.05}
\begin{tabular}{p{0.26\columnwidth}|p{0.25\columnwidth}|ccc}
\hline
\textbf{Slice} & \textbf{Primary metric} & \textbf{Baseline} & \textbf{Ours} & \textbf{$\Delta$} \\
\hline
Thin openings & Layout F1 @ IoU$_{2D}$@0.5 & 0.4023 & 0.4824 & +0.0801 \\
Cluttered furniture & Box F1 @ IoU$_{3D}$@0.25 & 0.4270 & 0.4798 & +0.0528 \\
Point sparsity / view coverage & Box F1 @ IoU$_{3D}$@0.25 & 0.0893 & 0.3043 & +0.2150 \\
\hline
\end{tabular}
\end{table}

\begin{figure}[!t]
\centerline{\includegraphics[width=0.5\textwidth, clip, trim=-0.5cm 3cm 12.5cm 1cm]{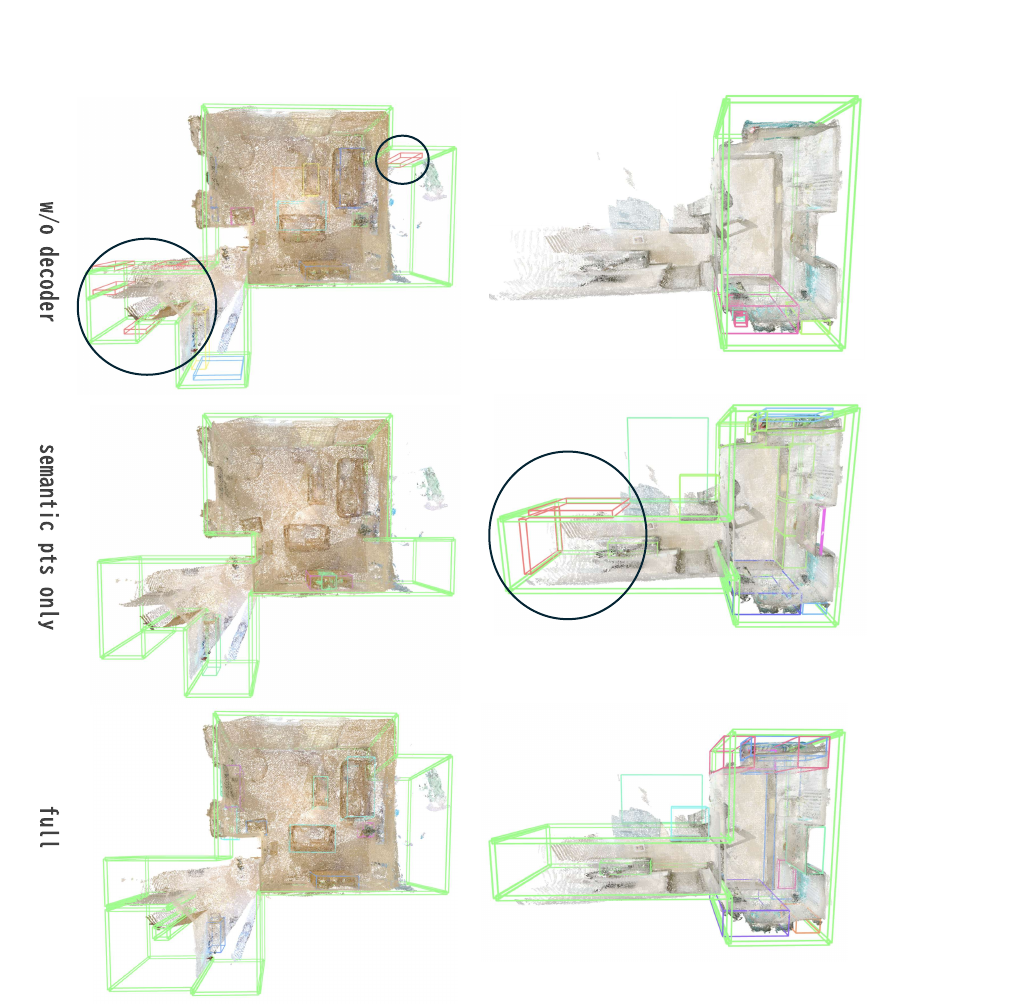}}
\caption{Qualitative comparison highlighting reduced empty-space hallucination. Compared to the baseline, our semantic-colored interface with shift suppresses spurious boxes in regions with weak or missing point support, while preserving accurate openings and furniture extents under sparse tokenization.}
\label{fig:abl1}
\end{figure}

\paragraph{Qualitative failure-mode reduction}
Figure~\ref{fig:abl1} complements Table~\ref{tab:failure_mode_slices} by showing that the proposed interface suppresses empty-space hallucinations instead of simply encouraging the decoder to emit more boxes.

\FloatBarrier

\section{Conclusion}
\label{sec:conclusion}
We presented an interface-preserving semantic-colored point conditioning mechanism for LLM-conditioned structured indoor prediction.
In the controlled setting, coarse semantic cues are lifted into a four-group RGBB code before tokenization so that semantics and geometry traverse the same sparse serialization/pooling path while the decoder-side protocol remains unchanged.
A minimal routed semantic shift together with lightweight ratio/budget/entropy constraints provides a constrained training and analysis design on top of this shared interface.

Across Structured3D, the SpatialLM dataset, and ARKitScenes, controlled point-cloud-interface evaluations show that the interface can improve opening localization and per-instance furniture box prediction when semantic support is well specified, while the click/SAM branch further shows that target-specific semantic-colored support on the aligned point input can bias box emission toward a selected instance without changing the downstream decoder.
The ablations also clarify the method's boundaries: gains are not uniform across all benchmarks and metrics, the auxiliary constraints mainly shape a layout--box trade-off, and the random four-color control shows that color/entropy confounds remain nontrivial.
The intensity-scaling study also indicates that this controllability is not smooth and comes with a nontrivial precision/false-positive trade-off.

Limitations therefore include the coarse semantic granularity, dependence on mask quality and view coverage, and the incomplete disentanglement between semantic meaning and color-channel effects. Future work includes richer yet robust semantic codes, confidence-aware multi-view fusion, and stronger controls for separating semantic signal from color/entropy confounds.
\bibliographystyle{elsarticle-harv}
\bibliography{eswa}

@inproceedings{schonberger2016structure,
  title={Structure-from-motion revisited},
  author={Schonberger, Johannes L and Frahm, Jan-Michael},
  booktitle={Proceedings of the IEEE conference on computer vision and pattern recognition},
  pages={4104--4113},
  year={2016}
}

@inproceedings{wang2024dust3r,
  title={{DUSt3R}: Geometric {3D} vision made easy},
  author={Wang, Shuzhe and Leroy, Vincent and Cabon, Yohann and Chidlovskii, Boris and Revaud, Jerome},
  booktitle={Proceedings of the IEEE/CVF Conference on Computer Vision and Pattern Recognition},
  pages={20697--20709},
  year={2024}
}

@inproceedings{leroy2024mast3r,
  title={Grounding image matching in {3D} with {MASt3R}},
  author={Leroy, Vincent and Cabon, Yohann and Revaud, J{\'e}r{\^o}me},
  booktitle={European Conference on Computer Vision},
  pages={71--91},
  year={2024},
  organization={Springer}
}

@inproceedings{wang2025vggt,
  title={{VGGT}: Visual geometry grounded transformer},
  author={Wang, Jianyuan and Chen, Minghao and Karaev, Nikita and Vedaldi, Andrea and Rupprecht, Christian and Novotny, David},
  booktitle={Proceedings of the Computer Vision and Pattern Recognition Conference},
  pages={5294--5306},
  year={2025}
}

@inproceedings{nie2020total3d,
  title={{Total3DUnderstanding}: Joint layout, object pose and mesh reconstruction for indoor scenes from a single image},
  author={Nie, Yinyu and Han, Xiaoguang and Guo, Shihui and Zheng, Yujian and Chang, Jian and Zhang, Jian Jun},
  booktitle={Proceedings of the IEEE/CVF Conference on Computer Vision and Pattern Recognition},
  pages={55--64},
  year={2020}
}

@inproceedings{murez2020atlas,
  title={Atlas: End-to-end {3D} scene reconstruction from posed images},
  author={Murez, Zak and Van As, Tarrence and Bartolozzi, James and Sinha, Ayan and Badrinarayanan, Vijay and Rabinovich, Andrew},
  booktitle={European conference on computer vision},
  pages={414--431},
  year={2020},
  organization={Springer}
}

@inproceedings{sun2021neuralrecon,
  title={{NeuralRecon}: Real-time coherent {3D} reconstruction from monocular video},
  author={Sun, Jiaming and Xie, Yiming and Chen, Linghao and Zhou, Xiaowei and Bao, Hujun},
  booktitle={Proceedings of the IEEE/CVF conference on computer vision and pattern recognition},
  pages={15598--15607},
  year={2021}
}

@inproceedings{yue2023roomformer,
  title={Connecting the dots: Floorplan reconstruction using two-level queries},
  author={Yue, Yuanwen and Kontogianni, Theodora and Schindler, Konrad and Engelmann, Francis},
  booktitle={Proceedings of the IEEE/CVF conference on computer vision and pattern recognition},
  pages={845--854},
  year={2023}
}

@inproceedings{kirillov2023sam,
  title={Segment anything},
  author={Kirillov, Alexander and Mintun, Eric and Ravi, Nikhila and Mao, Hanzi and Rolland, Chloe and Gustafson, Laura and Xiao, Tete and Whitehead, Spencer and Berg, Alexander C and Lo, Wan-Yen and others},
  booktitle={Proceedings of the IEEE/CVF international conference on computer vision},
  pages={4015--4026},
  year={2023}
}

@inproceedings{cheng2022mask2former,
  title={Masked-attention mask transformer for universal image segmentation},
  author={Cheng, Bowen and Misra, Ishan and Schwing, Alexander G and Kirillov, Alexander and Girdhar, Rohit},
  booktitle={Proceedings of the IEEE/CVF conference on computer vision and pattern recognition},
  pages={1290--1299},
  year={2022}
}

@inproceedings{peng2023openscene,
  title={{OpenScene}: {3D} scene understanding with open vocabularies},
  author={Peng, Songyou and Genova, Kyle and Jiang, Chiyu and Tagliasacchi, Andrea and Pollefeys, Marc and Funkhouser, Thomas and others},
  booktitle={Proceedings of the IEEE/CVF conference on computer vision and pattern recognition},
  pages={815--824},
  year={2023}
}

@article{takmaz2023openmask3d,
  title={{OpenMask3D}: Open-vocabulary {3D} instance segmentation},
  author={Takmaz, Ay{\c{c}}a and Fedele, Elisabetta and Sumner, Robert W and Pollefeys, Marc and Tombari, Federico and Engelmann, Francis},
  journal={arXiv preprint arXiv:2306.13631},
  year={2023}
}

@article{jatavallabhula2023conceptfusion,
  title={{ConceptFusion}: Open-set multimodal {3D} mapping},
  author={Jatavallabhula, Krishna Murthy and Kuwajerwala, Alihusein and Gu, Qiao and Omama, Mohd and Chen, Tao and Maalouf, Alaa and Li, Shuang and Iyer, Ganesh and Saryazdi, Soroush and Keetha, Nikhil and others},
  journal={arXiv preprint arXiv:2302.07241},
  year={2023}
}

@inproceedings{qi2017pointnet,
  title={PointNet: Deep learning on point sets for {3D} classification and segmentation},
  author={Qi, Charles R and Su, Hao and Mo, Kaichun and Guibas, Leonidas J},
  booktitle={Proceedings of the IEEE conference on computer vision and pattern recognition},
  pages={652--660},
  year={2017}
}

@inproceedings{qi2017pointnetplusplus,
  title={{PointNet++}: Deep hierarchical feature learning on point sets in a metric space},
  author={Qi, Charles Ruizhongtai and Yi, Li and Su, Hao and Guibas, Leonidas J},
  booktitle={Advances in Neural Information Processing Systems},
  volume={30},
  year={2017}
}

@inproceedings{thomas2019kpconv,
  title={{KPConv}: Flexible and deformable convolution for point clouds},
  author={Thomas, Hugues and Qi, Charles R and Deschaud, Jean-Emmanuel and Marcotegui, Beatriz and Goulette, Fran{\c{c}}ois and Guibas, Leonidas J},
  booktitle={Proceedings of the IEEE/CVF international conference on computer vision},
  pages={6411--6420},
  year={2019}
}

@inproceedings{zhao2021pointtransformer,
  title={Point transformer},
  author={Zhao, Hengshuang and Jiang, Li and Jia, Jiaya and Torr, Philip HS and Koltun, Vladlen},
  booktitle={Proceedings of the IEEE/CVF international conference on computer vision},
  pages={16259--16268},
  year={2021}
}

@inproceedings{qian2022pointnext,
  title={{PointNeXt}: Revisiting {PointNet++} with improved training and scaling strategies},
  author={Qian, Guocheng and Li, Yuchen and Peng, Houwen and Mai, Jinjie and Hammoud, Hasan and Elhoseiny, Mohamed and Ghanem, Bernard},
  booktitle={Advances in Neural Information Processing Systems},
  volume={35},
  pages={23192--23204},
  year={2022}
}

@inproceedings{yu2022point,
  title={{Point-BERT}: Pre-training {3D} point cloud transformers with masked point modeling},
  author={Yu, Xumin and Tang, Lulu and Rao, Yongming and Huang, Tiejun and Zhou, Jie and Lu, Jiwen},
  booktitle={Proceedings of the IEEE/CVF conference on computer vision and pattern recognition},
  pages={19313--19322},
  year={2022}
}

@inproceedings{pang2022pointmae,
  title={{Point-MAE}: Masked autoencoders for point cloud self-supervised learning},
  author={Pang, Yatian and Wang, Wenxiao and Tay, Francis E. H. and Liu, Wei and Tian, Yonghong and Yuan, Li},
  booktitle={European Conference on Computer Vision},
  pages={604--621},
  year={2022},
  organization={Springer},
  doi={10.1007/978-3-031-20086-1_35}
}

@inproceedings{wu2025sonata,
  title={{Sonata}: Self-supervised learning of reliable point representations},
  author={Wu, Xiaoyang and DeTone, Daniel and Frost, Duncan and Shen, Tianwei and Xie, Chris and Yang, Nan and Engel, Jakob and Newcombe, Richard and Zhao, Hengshuang and Straub, Julian},
  booktitle={Proceedings of the Computer Vision and Pattern Recognition Conference},
  pages={22193--22204},
  year={2025}
}

@article{mao2025spatiallm,
  title={{SpatialLM}: Training Large Language Models for Structured Indoor Modeling},
  author={Mao, Yongsen and Zhong, Junhao and Fang, Chuan and Zheng, Jia and Tang, Rui and Zhu, Hao and Tan, Ping and Zhou, Zihan},
  journal={arXiv preprint arXiv:2506.07491},
  year={2025},
  doi={10.48550/arXiv.2506.07491}
}

@article{baruch2021arkitscenes,
  title={{ARKitScenes}: A diverse real-world dataset for {3D} indoor scene understanding using mobile {RGB-D} data},
  author={Baruch, Gilad and Chen, Zhuoyuan and Dehghan, Afshin and Dimry, Tal and Feigin, Yuri and Fu, Peter and Gebauer, Thomas and Joffe, Brandon and Kurz, Daniel and Schwartz, Arik and others},
  journal={arXiv preprint arXiv:2111.08897},
  year={2021}
}

@article{wang2025pi,
  title={{\ensuremath{\pi^3}: Permutation-Equivariant Visual Geometry Learning}},
  author={Wang, Yifan and Zhou, Jianjun and Zhu, Haoyi and Chang, Wenzheng and Zhou, Yang and Li, Zizun and Chen, Junyi and Pang, Jiangmiao and Shen, Chunhua and He, Tong},
  journal={arXiv preprint arXiv:2507.13347},
  year={2025},
  doi={10.48550/arXiv.2507.13347}
}

@inproceedings{zheng2020structured3d,
  title={{Structured3D}: A large photo-realistic dataset for structured {3D} modeling},
  author={Zheng, Jia and Zhang, Junfei and Li, Jing and Tang, Rui and Gao, Shenghua and Zhou, Zihan},
  booktitle={European Conference on Computer Vision},
  pages={519--535},
  year={2020},
  organization={Springer}
}

@inproceedings{perez2018film,
  title={{FiLM}: Visual reasoning with a general conditioning layer},
  author={Perez, Ethan and Strub, Florian and De Vries, Harm and Dumoulin, Vincent and Courville, Aaron},
  booktitle={Proceedings of the AAAI Conference on Artificial Intelligence},
  volume={32},
  number={1},
  year={2018},
  doi={10.1609/aaai.v32i1.11671}
}

@inproceedings{de2017modulating,
  title={Modulating early visual processing by language},
  author={De Vries, Harm and Strub, Florian and Mary, J{\'e}r{\'e}mie and Larochelle, Hugo and Pietquin, Olivier and Courville, Aaron C},
  booktitle={Advances in Neural Information Processing Systems},
  volume={30},
  year={2017}
}

@inproceedings{avetisyan2024scenescript,
  title={{SceneScript}: Reconstructing scenes with an autoregressive structured language model},
  author={Avetisyan, Armen and Xie, Christopher and Howard-Jenkins, Henry and Yang, Tsun-Yi and Aroudj, Samir and Patra, Suvam and Zhang, Fuyang and Frost, Duncan and Holland, Luke and Orme, Campbell and others},
  booktitle={European Conference on Computer Vision},
  pages={247--263},
  year={2024},
  organization={Springer}
}

@article{carion2025sam,
  title={{SAM 3}: Segment anything with concepts},
  author={Carion, Nicolas and Gustafson, Laura and Hu, Yuan-Ting and Debnath, Shoubhik and Hu, Ronghang and Suris, Didac and Ryali, Chaitanya and Alwala, Kalyan Vasudev and Khedr, Haitham and Huang, Andrew and others},
  journal={arXiv preprint arXiv:2511.16719},
  year={2025},
  doi={10.48550/arXiv.2511.16719}
}

@article{wang2024qwen2,
  title={{Qwen2-VL}: Enhancing vision-language model's perception of the world at any resolution},
  author={Wang, Peng and Bai, Shuai and Tan, Sinan and Wang, Shijie and Fan, Zhihao and Bai, Jinze and Chen, Keqin and Liu, Xuejing and Wang, Jialin and Ge, Wenbin and others},
  journal={arXiv preprint arXiv:2409.12191},
  year={2024}
}

@article{zhang2026utonia,
  title={Utonia: Toward One Encoder for All Point Clouds},
  author={Zhang, Yujia and Wu, Xiaoyang and Yang, Yunhan and Fan, Xianzhe and Li, Han and Zhang, Yuechen and Huang, Zehao and Wang, Naiyan and Zhao, Hengshuang},
  journal={arXiv preprint arXiv:2603.03283},
  year={2026},
  doi={10.48550/arXiv.2603.03283}
}

@inproceedings{hedau2009recovering,
  title={Recovering the spatial layout of cluttered rooms},
  author={Hedau, Varsha and Hoiem, Derek and Forsyth, David},
  booktitle={2009 IEEE 12th International Conference on Computer Vision},
  pages={1849--1856},
  year={2009},
  organization={IEEE}
}

@inproceedings{lee2009geometric,
  title={Geometric reasoning for single image structure recovery},
  author={Lee, David C and Hebert, Martial and Kanade, Takeo},
  booktitle={2009 IEEE Conference on Computer Vision and Pattern Recognition},
  pages={2136--2143},
  year={2009},
  organization={IEEE}
}

@inproceedings{zou2018layoutnet,
  title={{LayoutNet}: Reconstructing the {3D} room layout from a single {RGB} image},
  author={Zou, Chuhang and Colburn, Alex and Shan, Qi and Hoiem, Derek},
  booktitle={Proceedings of the IEEE Conference on Computer Vision and Pattern Recognition},
  pages={2051--2059},
  year={2018},
  doi={10.1109/CVPR.2018.00219}
}

@inproceedings{sun2019horizonnet,
  title={{HorizonNet}: Learning room layout with {1D} representation and pano stretch data augmentation},
  author={Sun, Cheng and Hsiao, Chi-Wei and Sun, Min and Chen, Hwann-Tzong},
  booktitle={Proceedings of the IEEE/CVF Conference on Computer Vision and Pattern Recognition},
  pages={1047--1056},
  year={2019}
}

@inproceedings{hong20233dllm,
  title={{3D-LLM}: Injecting the {3D} world into large language models},
  author={Hong, Yining and Zhen, Haoyu and Chen, Peihao and Zheng, Shuhong and Du, Yilun and Chen, Zhenfang and Gan, Chuang},
  booktitle={Advances in Neural Information Processing Systems},
  volume={36},
  year={2023}
}

@inproceedings{qi2019votenet,
  title={{Deep Hough Voting} for {3D} object detection in point clouds},
  author={Qi, Charles R and Litany, Or and He, Kaiming and Guibas, Leonidas J},
  booktitle={Proceedings of the IEEE/CVF International Conference on Computer Vision},
  pages={9277--9286},
  year={2019}
}

@inproceedings{liu2021groupfree3d,
  title={Group-free {3D} object detection via transformers},
  author={Liu, Ze and Zhang, Zheng and Cao, Yue and Hu, Han and Tong, Xin},
  booktitle={Proceedings of the IEEE/CVF International Conference on Computer Vision},
  pages={2949--2958},
  year={2021}
}

@inproceedings{choy20194d,
  title={{4D} spatio-temporal convnets: Minkowski convolutional neural networks},
  author={Choy, Christopher and Gwak, JunYoung and Savarese, Silvio},
  booktitle={Proceedings of the IEEE/CVF Conference on Computer Vision and Pattern Recognition},
  pages={3075--3084},
  year={2019}
}

@inproceedings{long2015fcn,
  title={Fully convolutional networks for semantic segmentation},
  author={Long, Jonathan and Shelhamer, Evan and Darrell, Trevor},
  booktitle={Proceedings of the IEEE Conference on Computer Vision and Pattern Recognition},
  pages={3431--3440},
  year={2015},
  doi={10.1109/CVPR.2015.7298965}
}

@inproceedings{he2017maskrcnn,
  title={Mask {R-CNN}},
  author={He, Kaiming and Gkioxari, Georgia and Doll{\'a}r, Piotr and Girshick, Ross},
  booktitle={Proceedings of the IEEE International Conference on Computer Vision},
  pages={2961--2969},
  year={2017}
}

@inproceedings{cheng2021maskformer,
  title={Per-pixel classification is not all you need for semantic segmentation},
  author={Cheng, Bowen and Schwing, Alexander and Kirillov, Alexander},
  booktitle={Advances in Neural Information Processing Systems},
  volume={34},
  pages={17864--17875},
  year={2021}
}

@inproceedings{radford2021learning,
  title={Learning transferable visual models from natural language supervision},
  author={Radford, Alec and Kim, Jong Wook and Hallacy, Chris and Ramesh, Aditya and Goh, Gabriel and Agarwal, Sandhini and Sastry, Girish and Askell, Amanda and Mishkin, Pamela and Clark, Jack and Krueger, Gretchen and Sutskever, Ilya},
  booktitle={Proceedings of the 38th International Conference on Machine Learning},
  pages={8748--8763},
  year={2021},
  volume={139},
  series={Proceedings of Machine Learning Research},
  publisher={PMLR}
}

@inproceedings{kerr2023lerf,
  title={{LERF}: Language embedded radiance fields},
  author={Kerr, Justin and Kim, Chung Min and Goldberg, Ken and Kanazawa, Angjoo and Tancik, Matthew},
  booktitle={Proceedings of the IEEE/CVF International Conference on Computer Vision},
  pages={19729--19739},
  year={2023}
}

@inproceedings{shafiullah2023clipfields,
  title={{CLIP-Fields}: Weakly supervised semantic fields for robotic memory},
  author={Shafiullah, Nur Muhammad Mahi and Paxton, Chris and Pinto, Lerrel and Chintala, Soumith and Szlam, Arthur},
  booktitle={Proceedings of Robotics: Science and Systems},
  year={2023},
  address={Daegu, Republic of Korea},
  month={July},
  doi={10.15607/RSS.2023.XIX.074}
}

@inproceedings{jiang2023probabilistic,
  title={Probabilistic Triangulation for Uncalibrated Multi-View {3D} Human Pose Estimation},
  author={Jiang, Boyuan and Hu, Lei and Xia, Shihong},
  booktitle={Proceedings of the IEEE/CVF International Conference on Computer Vision},
  pages={14850--14860},
  year={2023}
}

@article{shuai2023adaptive,
  title={Adaptive Multi-View and Temporal Fusing Transformer for {3D} Human Pose Estimation},
  author={Shuai, Hui and Wu, Lele and Liu, Qingshan},
  journal={IEEE Transactions on Pattern Analysis and Machine Intelligence{ }},
  volume={45},
  number={4},
  pages={4122--4135},
  year={2023},
  doi={10.1109/TPAMI.2022.3188716}
}

@article{song2026ectformer,
  title={{ECTFormer}: Efficient {CNN}-Transformer Network for Uncalibrated Multiview {3-D} Human Pose Estimation},
  author={Song, Jucheng and Yang, Xu and Wang, Yapeng and Zhang, Jie and Im, Sio Kei},
  journal={IEEE Sensors Journal{ }},
  volume={26},
  number={6},
  pages={8487--8498},
  year={2026},
  doi={10.1109/JSEN.2026.3658078}
}

\end{document}